\documentclass[preprint,review,3p]{elsarticle}
\usepackage{hyperref}

\usepackage{amssymb}
\usepackage{caption}
\usepackage{subcaption}
\usepackage{multirow}

\journal{Journal of \LaTeX\ Templates}

\begin{document}

\begin{frontmatter}

\title{FPS-Net: A Convolutional Fusion Network\\ for Large-Scale LiDAR Point Cloud Segmentation}

\author[1]{Aoran Xiao}
\ead{aoran.xiao@ntu.edu.sg}

\author[2]{Xiaofei Yang}
\ead{xiaofei.hitsz@gmail.com}

\author[1]{Shijian Lu\corref{cor1}}%
\ead{Shijian.Lu@ntu.edu.sg}

\author[1]{Dayan Guan}
\ead{dayan.guan@ntu.edu.sg}

\author[1]{Jiaxing Huang}
\ead{jiaxing.huang@ntu.edu.sg}

\cortext[cor1]{Corresponding author}

\address[1]{Nanyang Technological University, 50 Nanyang Avenue, 639798, Singapore}
\address[2]{University of Macau, Avenida da Universidade
Taipa, Macau, 999078, China}

\begin{abstract}
Scene understanding based on LiDAR point cloud is an essential task for autonomous cars to drive safely, which often employs spherical projection to map 3D point cloud into multi-channel 2D images for semantic segmentation.
Most existing methods simply stack different point attributes/modalities (e.g. coordinates, intensity, depth, etc.) as image channels to increase information capacity, but ignore distinct characteristics of point attributes in different image channels. 
We design FPS-Net, a convolutional fusion network 
that exploits the uniqueness and discrepancy among the projected image channels for optimal point cloud segmentation. FPS-Net adopts an encoder-decoder structure. Instead of simply stacking multiple channel images as a single input, we group them into different modalities to first learn modality-specific features separately and then map the learnt features into a common high-dimensional feature space for pixel-level fusion and learning. Specifically, we design a residual dense block with multiple receptive fields as a building block in encoder which preserves detailed information in each modality and learns hierarchical modality-specific and fused features effectively. In the FPS-Net decoder, we use a recurrent convolution block likewise to hierarchically decode fused features into output space for pixel-level classification. Extensive experiments conducted on two widely adopted point cloud datasets show that FPS-Net achieves superior semantic segmentation as compared with state-of-the-art projection-based methods. In addition, the proposed modality fusion idea is compatible with typical projection-based methods and can be incorporated into them with consistent performance improvement.
\end{abstract}


\end{frontmatter}

\section{Introduction}\label{INTRODUCTION}

With the rapid development of LiDAR sensors and autonomous driving in recent years, LiDAR point cloud data which enable accurate measurement and representation of 3D scenes have been attracting increasing attention. Specifically, semantic segmentation of LiDAR point cloud data has become an essential function for fine-grained scene understanding by predicting a meaningful class label for each individual cloud point. On the other hand, semantic segmentation of large-scale LiDAR point cloud is still facing various problems. For example, LiDAR point cloud is heterogeneous by nature which captures both point geometries (with explicit point coordinates and depth) and laser intensity. In addition, it needs to store and process large-scale point cloud data collected within a limited time period, e.g., each scan in point cloud sequences may capture over 100k points. Further, the captured cloud points are often unevenly distributed in large spaces, where most areas have very sparse and even no points at all. Semantic segmentation of large-scale LiDAR point cloud remains a very open research challenge due to these problems.

A majority of recent works on point cloud segmentation make use of deep neural networks (DNNs) due to their powerful representation capability. As point cloud data are not grid-based with disordered structures, they cannot be directly processed by using deep convolution neural networks (CNNs) like 2D images. Several approaches have been proposed for DNN-based point cloud semantic segmentation \cite{guo2020deep}. For example, \cite{simonovsky2017dynamic,wang2019dynamic,bruna2013spectral} construct graphs by treating points as vertexes and extracts features via graph learning networks. \cite{qi2017pointnet,qi2017pointnet++,landrieu2018spg,tatarchenko2018tangent,hu2020randla} instead employ multi-layer perceptron (MLP) to learn from raw cloud points directly. However, all these works require neighbor search for building up neighborhood relations of points which is a computationally intensive process for large-scale point cloud \cite{liu2019point}. In addition, \cite{huang2016point,tchapmi2017segcloud,rethage2018fully} leverage 3D convolution to structurize cloud points, but they often face a dilemma of voxelization - it loses details if outputting low-resolution 3D grids but the computational costs and memory requirement increase cubically while voxel resolution increases.

In recent years, spherical projection has been exploited to map LiDAR sequential point cloud scans to depth images  and achieved superior segmentation \cite{wu2018squeezeseg,wu2019squeezesegv2,milioto2019rangenet++,shi2020spsequencenet}. As LiDAR point cloud is usually collected by linearly rotating scans, spherical projection provides an efficient way to sample and represent LiDAR point cloud scans in grid structure and the projected 2D images can be processed by developed CNNs in the similar way as regular 2D images. On the other hand, most existing spherical projection methods stack different point attributes (i.e. point coordinates x, y, and z, point intensity and point depth) as different channels in the projected images and process them straightly without considering their very diverse characteristics. We observe that the projected channel images actually have clear \textit{modality gaps}, and simply stacking them as regular images tends to produce sub-optimal segmentation. Specifically, we note that the projected channel images have three types of modality that have very different nature and distributions: The x, y, z record \textit{spatial coordinates} of cloud points in 3D Cartesian space, the \textit{depth} measures the distance between cloud points and LiDAR sensor in polar coordinate system, and the \textit{intensity} captures laser reflection. By simply stacking channel images of the three modalities as regular images, CNNs will process them with the same convolution kernels which tends to learn modality-agnostic common features but loses the very useful modality-specific features and information.

\begin{figure}[!h]
  \centering
  
  \begin{subfigure}[b]{\textwidth}
  \centering
  \includegraphics[width=\textwidth]{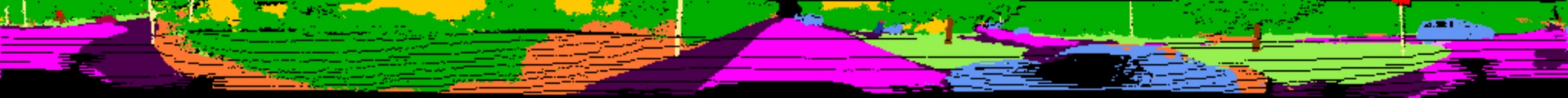}
  \caption{Prediction of \textit{depth} image}
  \end{subfigure}%
  
  \begin{subfigure}[b]{\textwidth}
  \centering
  \includegraphics[width=\textwidth]{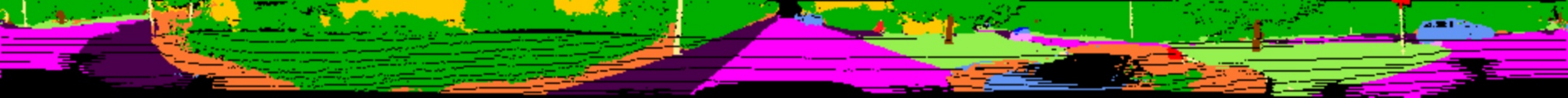}
  \caption{Prediction of \textit{coordinate} image}
  \end{subfigure}
  
  \begin{subfigure}[b]{\textwidth}
  \centering
  \includegraphics[width=\textwidth]{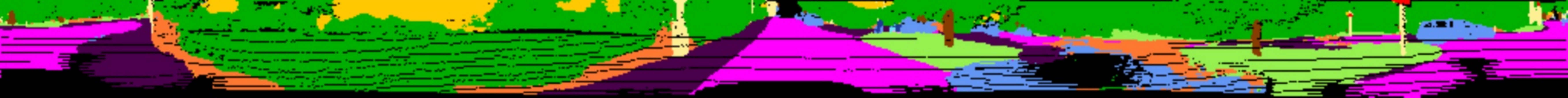}
  \caption{Prediction of \textit{intensity} image}
  \end{subfigure}
  
  \begin{subfigure}[b]{\textwidth}
  \centering
  \includegraphics[width=\textwidth]{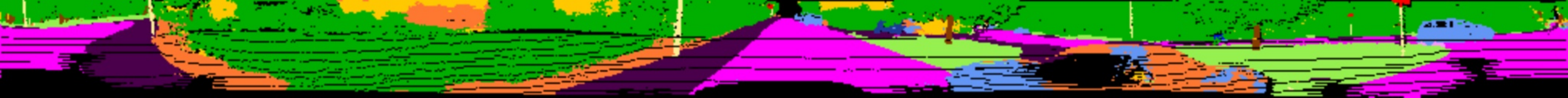}
  \caption{Prediction of the \textit{stacked} image with three modalities}
  \end{subfigure}
  
  \begin{subfigure}[b]{\textwidth}
  \centering
  \includegraphics[width=\textwidth]{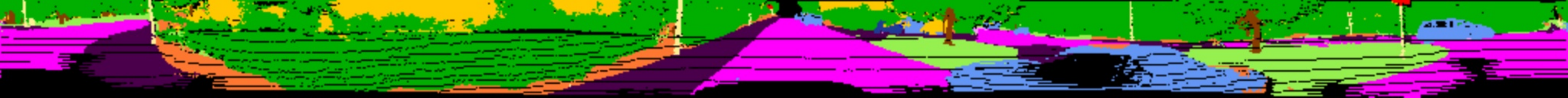}
  \caption{Prediction of FPS-Net}
  \end{subfigure}%
  
  \begin{subfigure}[b]{\textwidth}
  \centering
  \includegraphics[width=\textwidth]{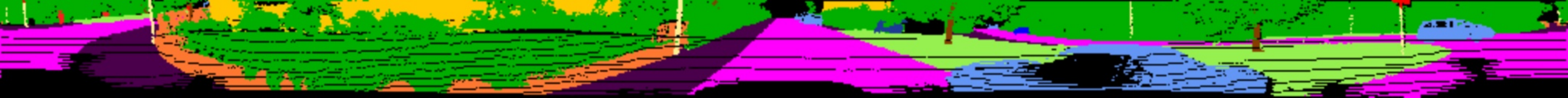}
  \caption{Ground Truth}
  \end{subfigure}
  
  \caption{Learning modality-specific features separately addresses modality gaps effectively: For a sample point cloud scan, (a), (b), and (c) show its segmentation by learning from the projected modality-specific depth images, coordinate (in x, y, and z) images, and intensity images, respectively. The segmentation is not accurate (e.g. in blue-color car that is close to the LiDAR sensor) but complementary across modalities. Stacking all channel images as one in learning does not improve segmentation clearly as in (d), large due to the negligence of modality gaps. Learning modality-specific features separately followed by fusion effectively mitigates the modality gap which improves the point cloud segmentation greatly as in (e).}
  \label{fig.PredCompare}
\end{figure}
  
We consider the modality gap and propose to learn modality-specific features from separate instead of stacked image channels. Specifically, we first learn features from modality-specific channel images separately and then map the learned modality features into a common high-dimensional feature space for fusion. We design an end-to-end trainable convolutional fusion network \textit{FPS-Net} that first learns features from each image modality and then fuses the learned modality features. Fig. \ref{fig.PredCompare} shows a sample scan that illustrates how FPS-Net addresses the modality gaps effectively. As the car does not get complete point cloud scan due to its proximity to LiDAR sensor, the predictions are very inconsistent across the three image modalities as illustrated in Figs. \ref{fig.PredCompare}(a), (b) and (c). Stacking all channels images as a single input does not improve the segmentation much as shown in Fig. 1(d) as it learns largely modality-agnostic features but misses modality-specific ones. As a comparison, FPS-Net produces much better segmentation as in Fig. 1(e) due to its very different design and learning strategy.

The contributions of this work can be summarized in three aspects. First, we identify an important but largely neglected problem in point cloud segmentation, i.e. modality gaps exist widely in the projected channel images which often leads to sub-optimal segmentation if unaddressed. Based on this observation, we propose to learn modality-specific features from the respective channel images separately instead of stacking all channel images as one in learning. Second, we design an end-to-end trainable modality fusion network that first learns modality-specific features and then maps them to a common high-dimensional space for fusion. Third, we conduct extensive experiments over two widely used public datasets, and experiments show that our network outperforms state-of-the-art methods clearly with similar computational complexity. Additionally, our idea can be straightly incorporated into most existing spherical projection methods with consistent performance improvements.

The rest of this paper is organized as follows. Section~\ref{Sec.relatework} reviews related works on point cloud semantic segmentation and modality fusion. Section \ref{Sec.methodology} describes the proposed method in details. Section \ref{Sec.experiments} presents experimental results and analysis and a few concluding remarks are finally drawn in Section \ref{Sec.conclusion}.

\section{Related Work}\label{Sec.relatework}
Our work is closely related to semantic segmentation of point cloud and multi-modality data fusion.

\subsection{Semantic Segmentation on Point Cloud}
We first introduce traditional point cloud segmentation methods briefly and then move to DNN-based methods including point-based methods and projection-based methods. We also introduce spherical projection methods due to their unique characteristics in large-scale LiDAR sequential point cloud segmentation.

\subsubsection{Traditional methods}
Extensive efforts have been made to the semantic segmentation of point cloud collected via laser scanning technology. 
Before the prevalence of deep learning, most works tackle point cloud segmentation with the following procedures \cite{xiang2019novel}:
1) unit/sample generation; 2) hand-craft feature design; 3) classifier selection; 4) post-processing. The first step aims to select a portion of points and generate processing units \cite{guo2011relevance,weinmann2014semantic,lehtomaki2015object,zhang2013svm,xu2014multiple}. Then discriminate features over units are extracted on feature design stage \cite{lehtomaki2015object,kumar2019feature,hackel2016fast}. Strong classifiers are chosen in the next stage to discriminate features of each unit\cite{zhang2013svm,zhou2012super}. Finally, post-processing such as conditional random fields (CRF) and graph cuts are implemented for better segmentation \cite{landrieu2017structured,niemeyer2013classification,guo2015classification}. 

\subsubsection{Deep learning methods}
Deep learning methods have demonstrated their superiority in automated feature extraction and optimization~\cite{lecun2015deep}, which also advanced the development of semantic segmentation of point cloud in recent years. One key to the success of deep learning is convolutional neural networks (CNN) that are very good at processing structured data such as images. 
However, CNNs do not perform well on semantic segmentation of point cloud data due to its unique characteristics in rich distortion, lack of structures, etc.  The existing deep learning methods can be broadly classified into two categories \cite{guo2020deep}, namely, point-based methods and projection-based methods.
Point-based methods process raw point clouds directly based on the fully connected neural network while projection-based methods firstly project unstructured point cloud into structured data and further process the structured projected data by using CNN models, 
more details to be reviewed in the ensuing two subsections.

\textit{Point-based methods:} A number of point-based methods have been reported which segment point cloud by directly processing raw point clouds without introducing any explicit information loss. One pioneer work is PointNet~\cite{qi2017pointnet} that extracts point-wise features from disordered points with shared MLPs and symmetrical pooling. Several ensuing works ~\cite{qi2017pointnet++,choy20194d,zhao2019pointweb,zhang2019shellnet} aim to improve PointNet from different aspects. For example, ~\cite{yang2019modeling,zhao2019pooling} introduce attention mechanism ~\cite{vaswani2017attention} for capturing more local information. In addition, a few works~\cite{tatarchenko2018tangent,hua2018pointwise,thomas2019kpconv} build customized 3D convolution operations over unstructured and disordered cloud points directly. For example, ~\cite{tatarchenko2018tangent} implements tangent convolution on a virtual tangent surface from surface geometry of each point. \cite{hua2018pointwise} presents a point-wise convolution network that groups neighbouring points into cells for convolution. \cite{thomas2019kpconv} introduces Kernel Point Convolution for point cloud processing. 

Though the aforementioned methods achieved quite impressive accuracy, most of them require heavy computational costs and large network models with a huge number of parameters. These requirements are unacceptable for many applications that require instant processing of tens of thousands of point in each scan as collected by high-speed LiDAR sensors (around 5-20 Hz). To mitigate these issues, ~\cite{hu2020randla} presents a lightweight but efficient network RandLA-Net that employs random sampling for real-time segmentation of large-scale LiDAR point cloud. ~\cite{zhang12356deep} presents a voxel-based “mini-PointNet” that greatly reduces memory and computational costs. ~\cite{tang2020searching} presents a 3D neural architecture search technique that searches for optimal network structures and achieves very impressive efficiency.

\textit{Projection-based methods:} 
Projection-based methods project 3D point cloud to structured images for processing by well-developed CNNs. Different projection approaches have been proposed. \cite{lawin2017deep} presents a \textit{multiview representation} that first projects 3D point cloud to 2D planes from different perspectives and then applies multi-stream convolution to process each view. To preserve the sacrificed geometry information, ~\cite{huang2016point,tchapmi2017segcloud,meng2019vv} presents a \textit{volumetric representation} that splits 3D space into structured voxels and applies 3D voxel-wise convolution for semantic segmentation. However, the \textit{volumetric representation} is computationally prohibitive at high resolutions but it loses details and introduces discretization artifacts while working at low resolutions. To address these issues, SPLATNet~\cite{su2018splatnet} introduces the \textit{permutohedral lattice representation} where permutohedral sparse lattices are interpolated from raw points for bilateral convolution and the output is then interpolated back to the raw point cloud. Further, ~\cite{zhang2020polarnet} introduces PolarNet that projects point cloud into bird-eye-view images and quantifies points into polar coordinates for processing.

Beyond the aforementioned methods, \textit{spherical projection} ~\cite{wu2018squeezeseg} has been widely adopted for efficient large-scale LiDAR sequential point cloud segmentation. Specifically, spherical projection provides an efficient way to sample points and the projected images preserve geometric information of point cloud well and can be processed by standard CNNs effectively. A series of methods adopted this approach~\cite{wang2018pointseg, wu2018squeezeseg, wu2019squeezesegv2, milioto2019rangenet++, xu2020squeezesegv3,cortinhal2020salsanext,alonso20203d}. For example, ~\cite{wu2018squeezeseg,wang2018pointseg,wu2019squeezesegv2} employ SqueezeNet~\cite{iandola2016squeezenet} as backbone and Conditional Random Field (CRF) for post-processing. To enhance the scalability of spherical project in handling more and fine-grained classes, ~\cite{milioto2019rangenet++} presents an improved U-Net model \cite{ronneberger2015u} and GPU-accelerated post-processing for back-projecting 2D prediction to 3D point cloud. However, all these methods simply stack point data of different modalities (coordinate, depth, and intensity) as inputs without considering their heterogeneous distributions. We observe that the three modalities of point data have clear gaps and propose a separate learning and fusing strategy to mitigate the modality gap for optimal point cloud segmentation.

\subsection{Multi-Modality Data Fusion}
Modality refers to a certain type of information or the representation format of stored information. Multimodal learning is intuitively appealing as it could learn richer and more robust representations from multimodal data.
On the other hand, effective learning from multimodal data is nontrivial as multimodal data usually have different statistical distributions and highly nonlinear relations across modalities \cite{liu2018learn}.
Multimodal learning has been investigated in different applications. For example, \cite{liao2014generalized,ghamisi2015land,rasti2017hyperspectral,rasti2017fusion,hong2020more} fuse hyper-spectral images and LiDAR data for land-cover classification. \cite{dechesne2017semantic,sun2018developing,hong2020more} fuse multi-spectral images with LiDAR data for land-cover and land-use classification. \cite{qi20173d} learns visual information from visible RGB images and geometric information from point cloud for indoor semantic segmentation. \cite{guan2019fusion,cao2019box} fuses visible and thermal images for pedestrian detection.

In addition, different fusion techniques have been proposed~\cite{atrey2010multimodal,liu2018learn}, including early and late fusion~\cite{snoek2005early,gunes2005affect}, hybrid fusion~\cite{bendjebbour2001multisensor,xu2006fusion}, joint training \cite{ngiam2011multimodal,sun2019not}, and multiplicative multi-modal method~\cite{liu2018learn}. Different from these works that fuse multimodal data from different sensors, we will study multimodal data in LiDAR point cloud including coordinates (x,y,z), intensity and depth which are usually well aligned as they are produced from the same LiDAR sensor laser signal.

\section{Methodology}\label{Sec.methodology}

\subsection{Modality Gap and Proposed Solution}\label{Sec.problem_definition}
This paper focuses on semantic segmentation of LiDAR sequential point cloud. 
Given scans of LiDAR  point cloud with corresponding $C$ classes point-level labels in the training set and the testing set, our goal is to learn a point cloud semantic segmentation model $F_{3D}$ on training set that also performs well on the testing set. 

Previous spherical projection-based methods follow three steps in the pipeline: For $i_{th}$ scan of LiDAR point cloud $S_i\subset \mathbb{R}^{N_i \times 4}$ (with $N_i$ points and each point has four attributes values \textit{x, y, z, intensity}) in the training set : 1) A projection approach serving as a mapping function $\Phi$ transfers point cloud from 3D space to a 2D projected image $I_{i}=\Phi(S_i)$; 2) A convolutional neural network $F_{2D}$ extracts features from the image and predicts class distribution for all pixels,  i.e. $\hat{Y_i}=F_{2D}(I_{i})$. 3) A Post-processing function $\Psi$ maps 2D image predictions back to 3D space and outputs final prediction for all points $\hat{L}_i=\Psi(\hat{Y_i})$. The purpose is to minimize distance $D$ between 3D prediction $\hat{L}_i$ and its ground truth $L_i$:
\begin{equation}
  D(\hat{L}_i, L_i)=D(\Psi(F_{2D}(\Phi(S_i))), L_i)
\end{equation}

\begin{figure}
  \begin{subfigure}[b]{0.3\textwidth}
  \includegraphics[width=\textwidth]{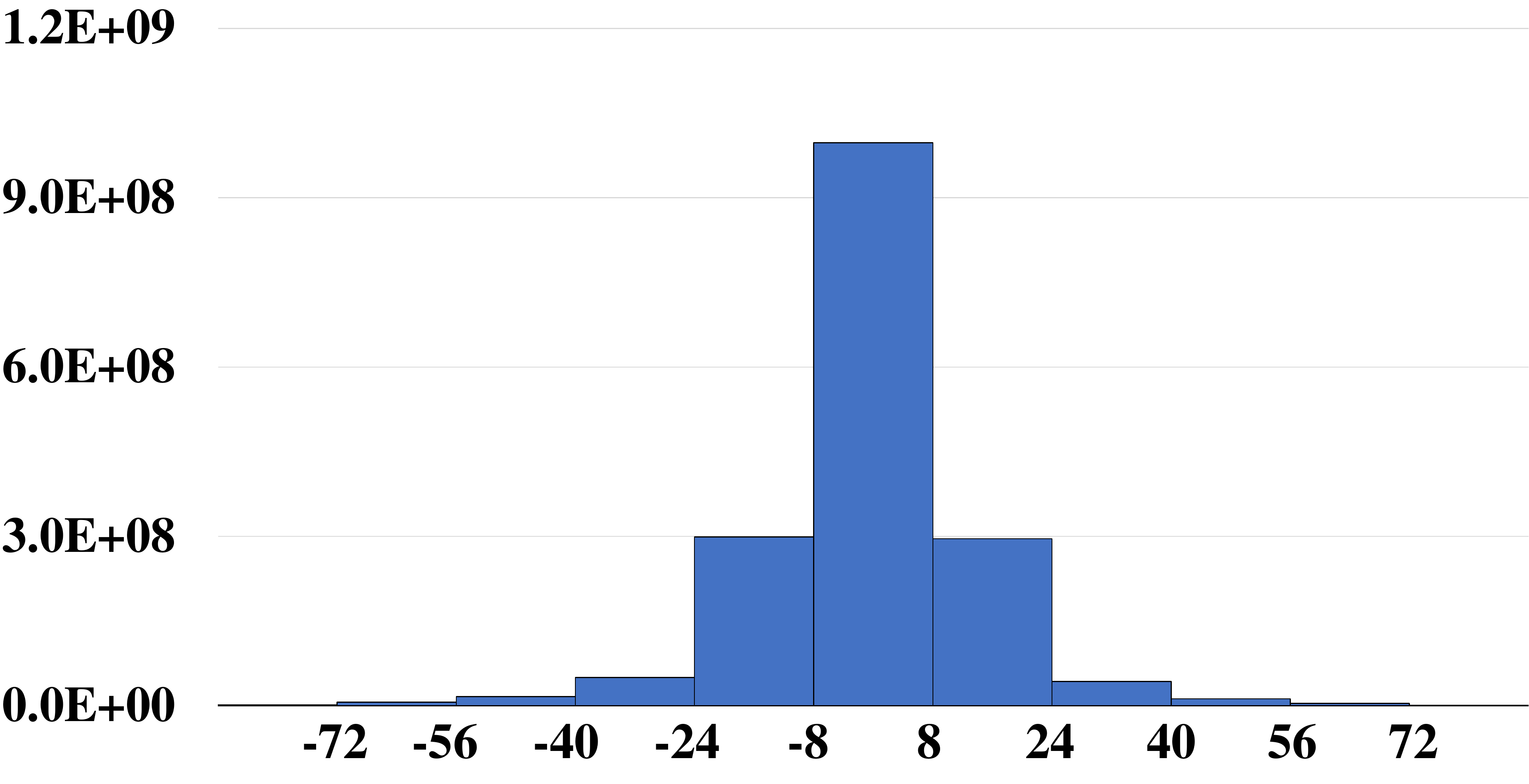}
  \caption{X}
  \end{subfigure}
  \begin{subfigure}[b]{0.3\textwidth}
  \includegraphics[width=\textwidth]{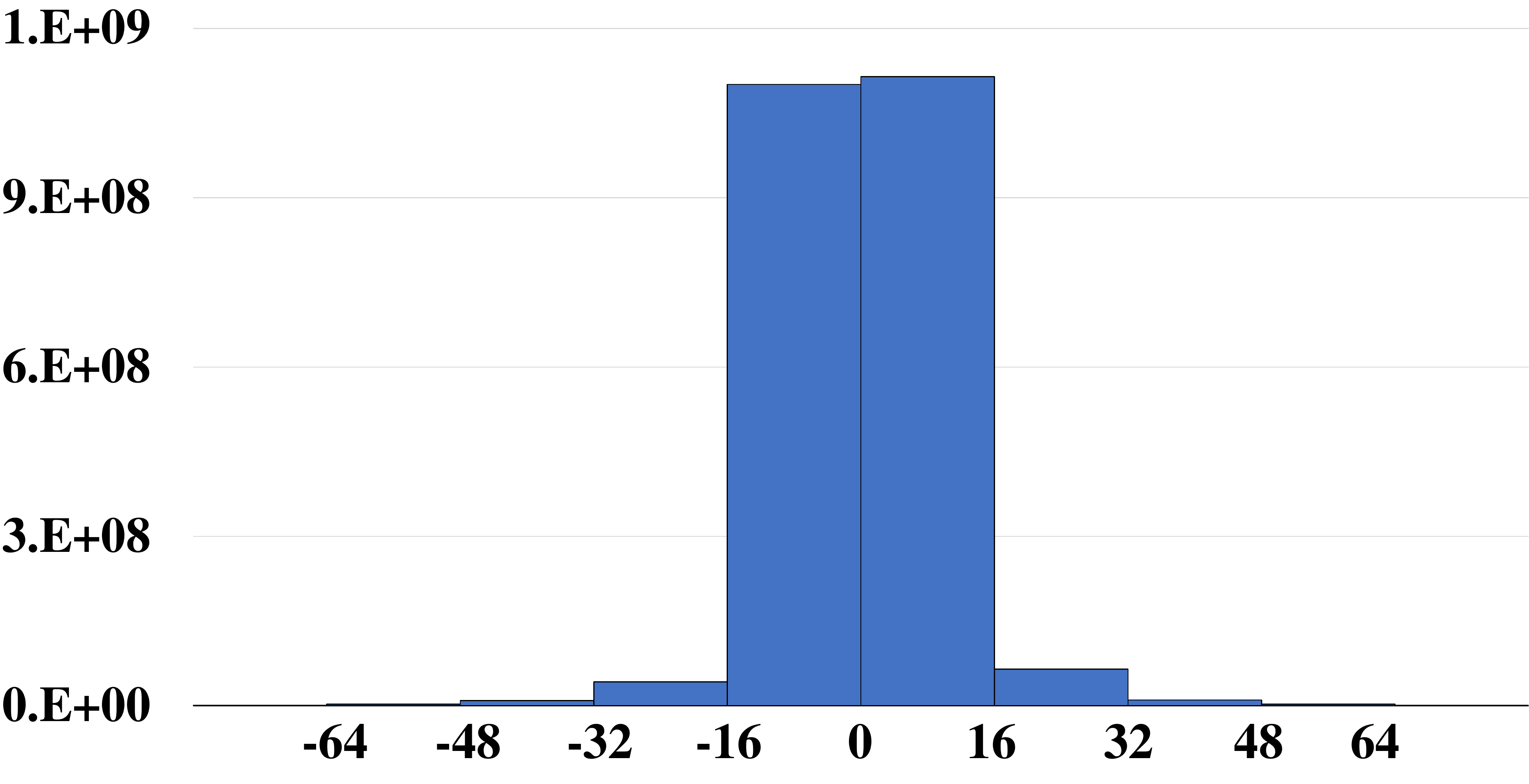}
  \caption{Y}
  \end{subfigure}
  \begin{subfigure}[b]{0.3\textwidth}
  \includegraphics[width=\textwidth]{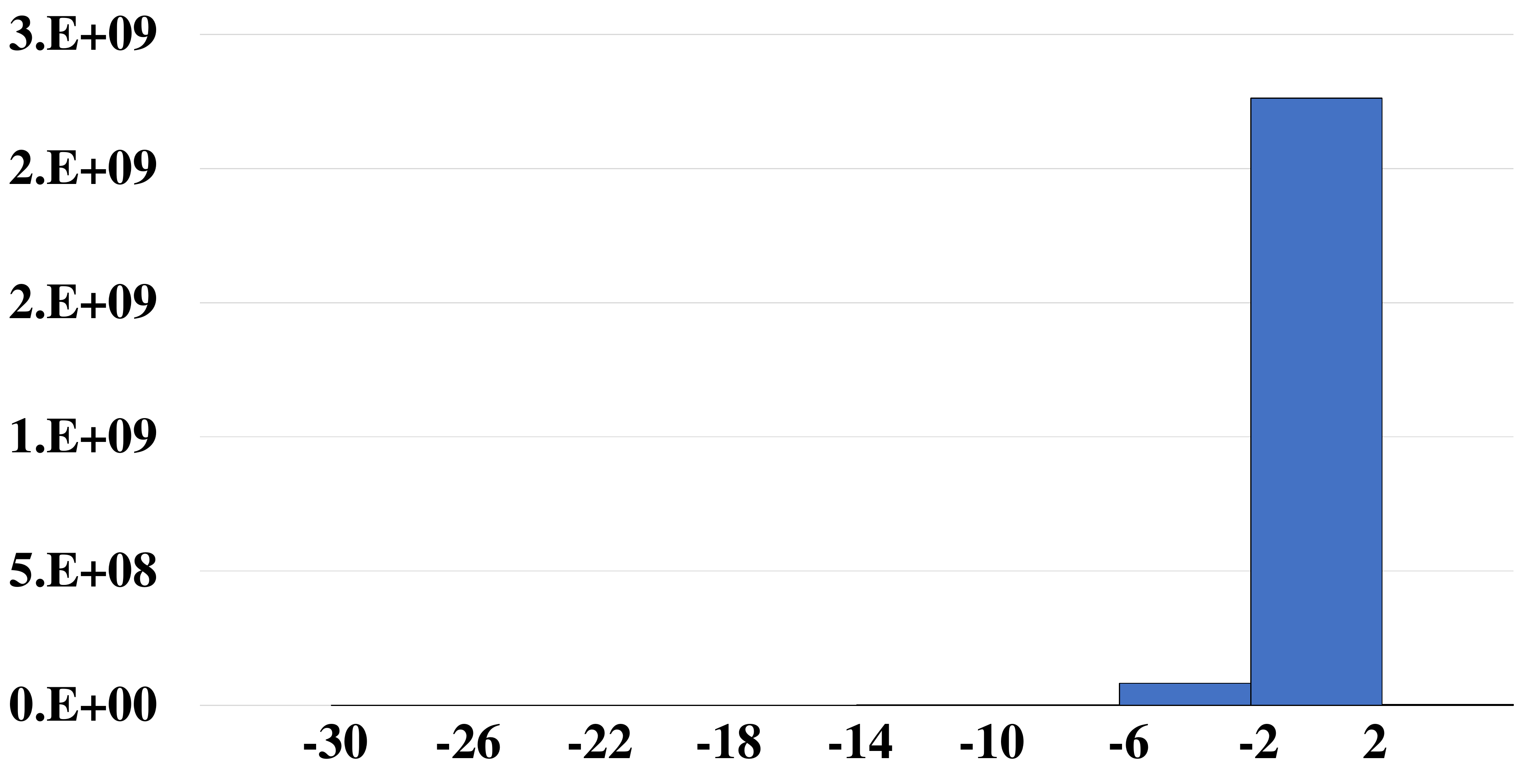}
  \caption{Z}
  \end{subfigure}
  
  \begin{subfigure}[b]{0.3\textwidth}
  \includegraphics[width=\textwidth]{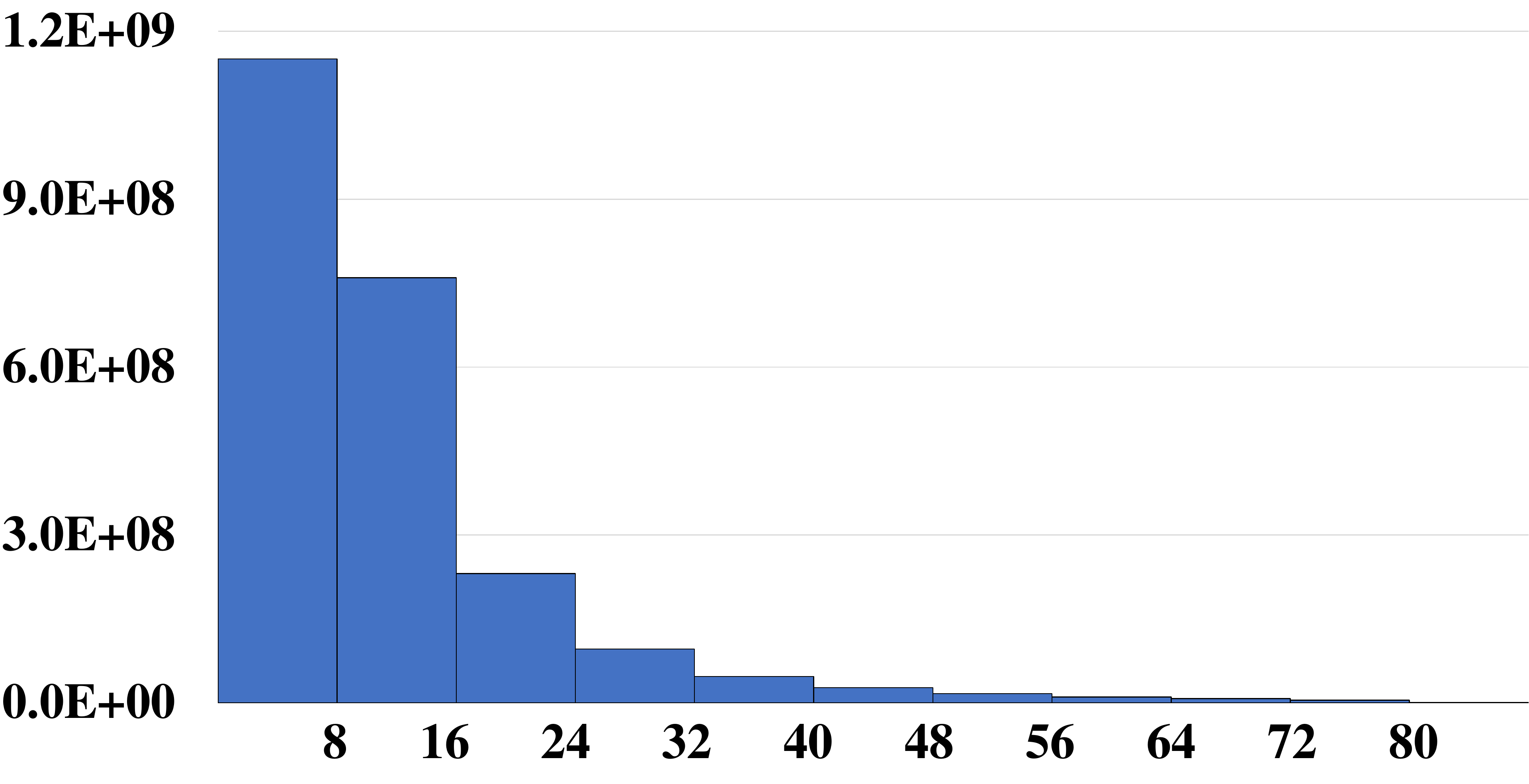}
  \caption{depth}
  \end{subfigure}
  \begin{subfigure}[b]{0.3\textwidth}
  \includegraphics[width=\textwidth]{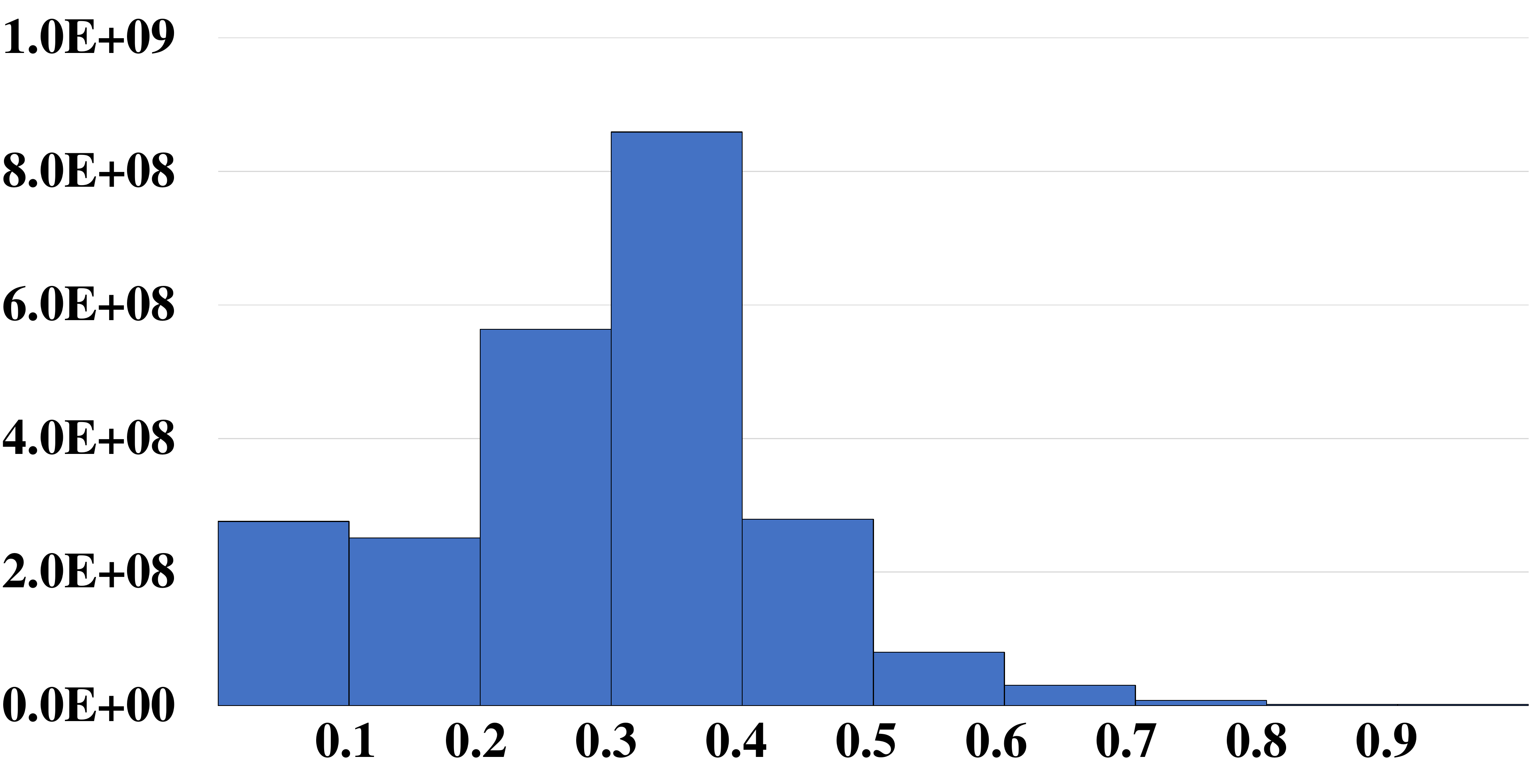}
  \caption{intensity}
  \end{subfigure}

  \caption{The distribution of 5 spherically projected channel images over the training set of SemanticKITTI, X-axis represents values for each modality and Y-axis shows corresponding number of pixels: The 5 projected channel images in modalities coordinate (x,y,z), depth and intensity have very different distribution. Simply stacking them together as one input in training ignores the modality differences, which often leads to sub-optimal model and semantic segmentation.}\label{fig.DistVary}
\end{figure}

Following prior work by \cite{wu2018squeezeseg}, existing methods \cite{wu2018squeezeseg,wu2019squeezesegv2,xu2020squeezesegv3, milioto2019rangenet++, alonso20203d, cortinhal2020salsanext} stack coordinates (x,y,z), depths and intensities as five channels in the projected images, i.e. $I_{i}\subset \mathbb{R}^{H \times W \times 5}$. Stacking different attribute values is able to increase informative capacity of input images but also brings modality gap problem. By counting on all pixel values in the training set of SemanticKITTI~\cite{behley2019semantickitti}, as Fig. \ref{fig.DistVary} shows, it is obvious that these three modalities follow different distributions. 
Simply stacking values following various distribution in the same image is likely to mislead models to focus on modality-agnostic features while ignore modality-specific information or even downgrade representation of the model to distinguish samples lie in distribution boundary spaces, leading to a sub-optimal segmentation result. 

Our solution is dividing each modality into an individual image for separate learning, in this case includes coordinate image $I_{D_i}$,  depth image $I_{C_i}$, and intensity image $I_{V_i}$,
\begin{equation}
\left\{ 
  \begin{array}{lr}
  I_{D_i}=\Phi_{D}(S_i) \\
  I_{C_i}=\Phi_{C}(S_i) \\
  I_{V_i}=\Phi_{V}(S_i) \\
  \end{array}
\right.
\end{equation}
and mapping them into a common high-dimensional space for fusion and further learning. The fusion approach is taken as the following equation in this shared data space: 
\begin{equation}
\hat{T}_{i}=[\hat{T}_{D_i}, \hat{T}_{C_i}, \hat{T}_{V_i}], where
\left\{ 
  \begin{array}{lr}
  \hat{T}_{D_i}=F_{D}(I_{D_i}) \\
  \hat{T}_{C_i}=F_{C}(I_{C_i}) \\
  \hat{T}_{V_i}=F_{V}(I_{V_i}) \\
  \end{array}
\right.
\end{equation}
$[\cdot]$ represents pixel-wise fusion. Then the 2D convolution neural network $F_{2D}$ learns from fused features to further predict semantic segmentation results, i.e. $\hat{Y_i}=F_{2D}(\hat{T}_{i})$. 

Overall, the distance could be summarize as
\begin{equation}
D(\hat{L}_i, L_i)=D(\Psi(F_{2D}([F_{D}(I_{D_i}), F_{C}(I_{C_i}), F_{V}(I_{V_i})])), L_i)
\end{equation}

Based on this idea, we design an end-to-end trainable modality fusion network FPS-Net that fully learns hierarchical information from each individual modality as well as the fused feature space. 

\subsection{Modality-Fused Convolution Network}

\begin{figure}
  \centering
  \includegraphics[width=1\textwidth]{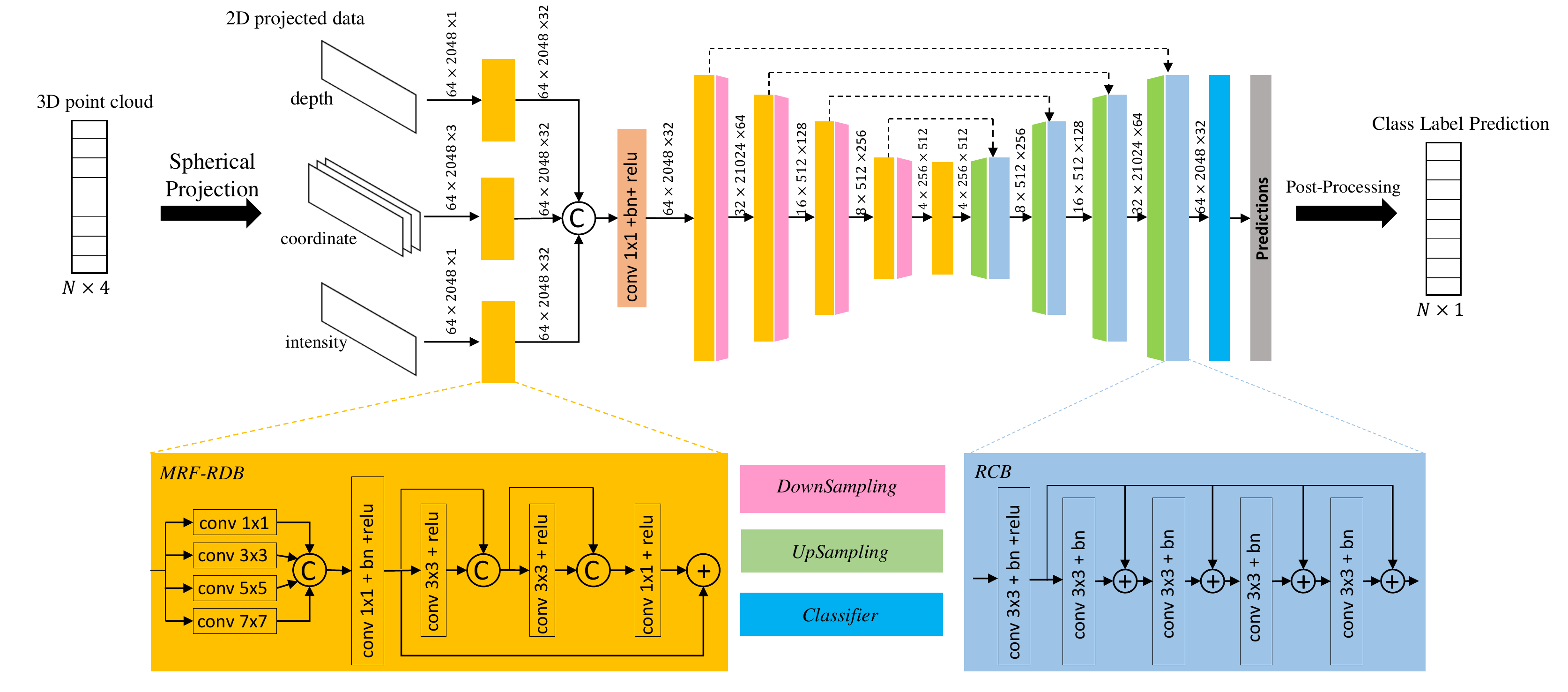}
  \caption{The architecture of our proposed FPS-Net : A spherical projection maps each point cloud scan into three modality images. The fusion learning network learns features of each modality separately and fuses the learned modality-specific features in a common high-dimensional space. After encoding and decoding the fused feature, FPS-Net maps 2D predictions back into 3D space for point-level semantic prediction. MRF-RDB and RCB are building blocks of encoder and decoder respectively for learning hierarchical features. Best viewed in color.}
  \label{fig.model}
\end{figure}

The whole architecture of the FPS-Net model is illustrated as Fig. \ref{fig.model}. 
A detailed introduction of each part is demonstrated in rest of this section.

\subsubsection{Mapping From 3D to 2D} 

The widely-used spherical projection~\cite{wu2018squeezeseg,wu2019squeezesegv2, milioto2019rangenet++,xu2020squeezesegv3,alonso20203d,cortinhal2020salsanext,shi2020spsequencenet} transfers 3D point cloud into 2D images through following equation $\Phi$. The pixel coordinate $(u,v)$ of a point with spatial coordinate $(x,y,z)$ in a single scan of point cloud could be calculated as:
\begin{equation}
  \left\{
    \begin{array}{lr}
    u=\frac{1}{2}[1-\arctan(y,x)\pi^{-1}]W & \\
    v=[1-(\arcsin(z, r^{-1})+f_{up})f^{-1}]H
    \end{array}
  \right.
\end{equation}
where $r$ is depth ($r=\sqrt{x^2+y^2+z^2}$); $f_{down}$ and $f_{up}$ represent lower and upper boundaries of vertical field-of-view of LiDAR sensor ($f=f_{up}+f_{down}$); $(H, W)$ are height and width of projected images. 

\begin{figure}[ht]
  \centering
  \begin{subfigure}[b]{0.5\textwidth}
    \includegraphics[width=\textwidth]{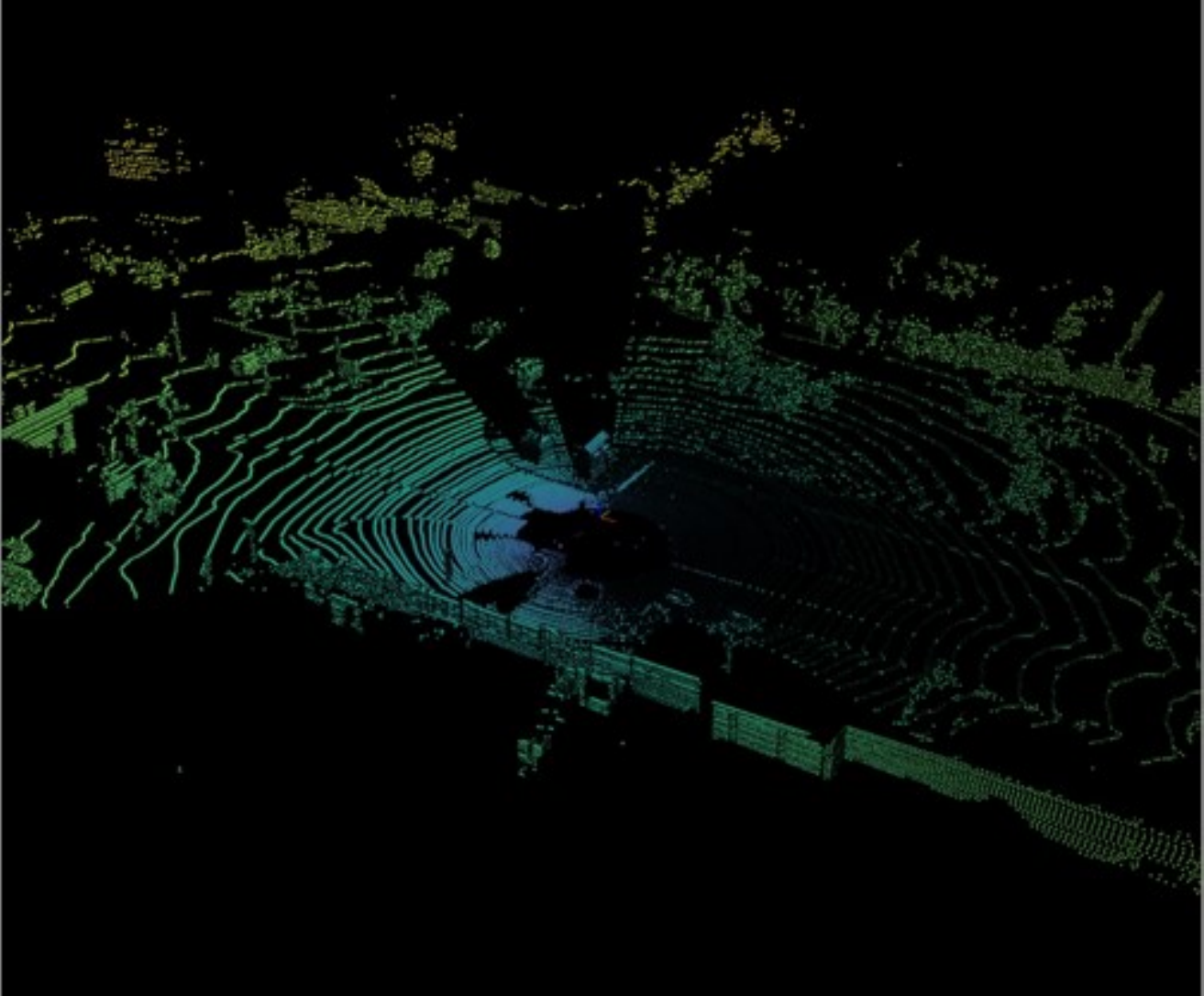}
    \caption{visualization of 3D LiDAR point cloud}
  \end{subfigure}
  \hfill
  \begin{subfigure}[b]{\textwidth}
    \includegraphics[width=\textwidth]{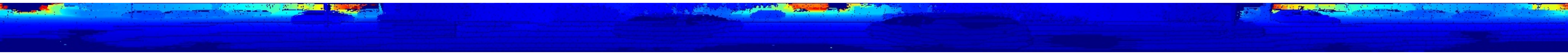}
    \caption{projected depth image}
  \end{subfigure}
  \hfill
  \begin{subfigure}[b]{\textwidth}
    \includegraphics[width=\textwidth]{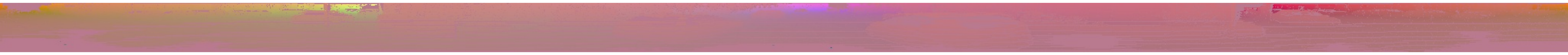}
    \caption{projected coordinate image}
  \end{subfigure}
  \hfill
  \begin{subfigure}[b]{\textwidth}
    \includegraphics[width=\textwidth]{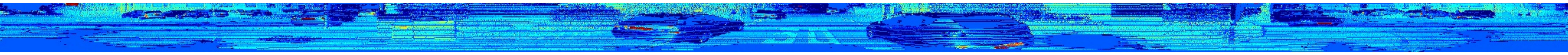}
    \caption{projected intensity image}
  \end{subfigure}
  \caption{Example of spherical projection from 3D point cloud space to 2D images. (a) a scan of point cloud; (b), (c), (d) are individual modality images.}\label{figure.projection}
\end{figure}

Instead of using stacked five-channel images as input as described in Section \ref{Sec.problem_definition}, we address the modality gaps by dividing the five-channel images into three modalities in point coordinates, point depths and point intensities. Fig. \ref{figure.projection} shows an example of point cloud and its corresponding modality images. Note each pixel in the coordinate image has three channel values (x, y, z), i.e. $I_{C_i}\subset \mathbb{R}^{H \times W \times 3}$, whereas the pixel in the depth image $I_{D_i}\subset \mathbb{R}^{H \times W \times 1}$ and intensity image $I_{V_i}\subset \mathbb{R}^{H \times W \times 1}$ has a single depth value and a single intensity value, respectively.

\subsubsection{Modality Fusion Learning}\label{Sec.domain_fusion_learn}
FPS-Net adopts a well designed encoder-decoder neural network to learn modality-specific features  and fused feature for semantic segmentation. 
As Fig. \ref{fig.model} illustrates, at the beginning of the model, three branches of feature extractors are designed to learn detailed information of each input modality separately. In order to capture rich information of individual images, we use a multiple receptive field residual dense block (MRF-RDB) as basic building block of feature extractors. The structure of MRF-RDB is demonstrated in Fig. \ref{fig.model}. A multi-receptive fields module is introduced at beginning. Four kernels with small to large sizes are in parallel for several considerations: Firstly, different from regular images, projected images usually contain many blank pixels due to lack of remission. A larger receptive field could capture more pixels to avoid insufficient learning. Secondly, such a parallel structure is able to capture objects in different sizes for better feature representation. Thirdly, we only use large receptive fields in the beginning of the building block to keep a smaller model size. Considering that the height of projected images is small, usually set as the number of raiser beams of LiDAR sensors (e.g. $H=64$ for Velodyne HDL-64E sensor), small object features are likely to be lost in deep layers, hence we adopt the dense residual layers~\cite{zhang2018residual} in the remaining part of MRF-RDB to learn hierarchical features. Specifically, it comprises multiple convolution layers with batchnorm layers and activation layers. The output of each layer then concatenates previous output features before forwarding to next layer, leading to a dense structure with a contiguous memory mechanism. Such structure adaptively captures and memorizes effective information from preceding to current local features. The output of the last convolution layer add the input of RDB for another hierarchical feature aggregation. 

After learning modality-specific high dimensional features, we fuse them by concatenation operation following a $1\times1$ convolution for dimension reduction. 
The fused feature is then sent into an sub-network as U-Net~\cite{ronneberger2015u} structure for further learning. The encoder of this sub-network consists of four MRF-RDB blocks and downsampling layers. The bridge between the encoder and the decoder is also a MRF-RDB but its output has the same dimension as input feature map, which is used to convey deep abstract signals into the decoder. 
We no longer use MRF-RDB in the decoder because after passing 4 MRF-RDB of encoder the fused feature is well extracted and a simpler decoder structure is better for classification. Instead, our previous design~\cite{yang2019road} of recurrent convolution block (RCB) with upsampling layer is used as the building block. As Fig. \ref{fig.model} shows, RCB also comprises several convolution layers. It integrates low-level and high-level features via addition instead of concatenation, which reduces parameters without dropping performances while keeps large receptive fields and preserves more detailed spatial information. We compared different combinations of these two building blocks in ablation study  (Section \ref{Sec.ablation_study}) and the result proved our idea. The classifier for 2D image prediction is a $3\times3$ convolution kernel with a softmax layer to predict probabilities of label classes for each pixel. The output channel sizes for five MRF-RDBs are 32, 64, 128, 256, 512 respectively, while the output channels for RCBs in decoder are 256, 128, 64, 32. 

The last step is mapping 2D prediction back to 3D space and assigning labels for each point. This step is ill-posed as some points may not be selected on images in the spherical projection step, especially when image size is small. This paper follows the post-process of RangeNet++~\cite{milioto2019rangenet++} and uses a GPU-supported KNN procedure to assign labels for missing points. In spherical projection step, each point $p_k$ is projected into a pixel coordinate $(u_k, v_k)$ (although only the nearest point would be selected as pixel when multiple points fall on the same pixel). In post-process, we use a $S\times S$ window centered on $(u_k, v_k)$ to get $S^2$ neighbor pixels of the projected image as candidates. For those $K_k$ points whose depth distance is smaller than a threshold $\sigma$, labels are voted from them with weight from distance by an inverse Gaussian kernel, which penalizes points with larger distances. More details could be found in \cite{milioto2019rangenet++}.

\subsubsection{Optimization}
Due to the uneven distribution of ground objects in wild,  LiDAR point cloud datasets~\cite{behley2019semantickitti,geiger2012kitti} often face a serious data-imbalanced problem, which negatively leads the model to focus on classes with more samples. 
Take SemanticKITTI dataset \cite{behley2019semantickitti} as example, classes including \textit{vegetation, road, building} and so on take majority of data while classes like \textit{bicycle, motorcycle} have very limited sample points.
Hence, we adopt the weighted softmax cross-entropy objective to focus more on classes with less samples, in which the weight of loss for each class is inverse to its frequency $f$ on training set:
\begin{equation}
    \mathcal{L}_{xent}=-\sum^{C}_{j=1}w_jp(y_j)\log(p(\hat{y}_j)), \ \ w_j=f_j^{-1}
\end{equation}
where $y_j$ and $\hat{y}_j$ are ground truth label and predicted label of class $j$. In this case, classes with small number of samples have larger losses and imbalanced data distribution problem could be alleviated.

We also adopt \textit{Lov\'{a}sz-Softmax} loss~\cite{berman2018lovasz} used in image semantic segmentation to improve learning performances, which regards IoU as a hypercube and its vertices are the combination of each class labels. In this case, the interior area of hypercube could be considered as IoU. The \textit{Lov\'{a}sz-Softmax} loss is defined as
\begin{equation}
\mathcal{L}_{lvsz}=\frac{1}{C}\sum_{j\in C}
\bar{\Delta_{J_j}}(l(j)) \\
\space l(j)=\left \{
    \begin{array}{ll}
         & 1-s(j), if j=y(j) \\
         & s(j),  else
    \end{array}
    \right.
\end{equation}
where $\bar{\Delta_{J_j}}$ is the \textit{Lov\'{a}sz} extention of Jaccard index. $s(j)$ is probability for pixel belonging to class $j$ and $y(j)$ is corresponding ground truth class label. 

The final objective of FPS-Net is summarized as $\mathcal{L}=\mathcal{L}_{xent}+\lambda\mathcal{L}_{lvsz}$. And the optimization process for training set could be illustrated as 
\begin{equation}
  F_{3D}=\mathop{\arg\min}\limits_{F_{3D}}\mathcal{L}=\mathop{\arg\min}\limits_{F_{3D}}(\mathcal{L}_{xent}+\lambda\mathcal{L}_{lvsz})
\end{equation}

\section{Experiments}\label{Sec.experiments}
\subsection{Experimental Setup}
\textbf{Datasets:}
We evaluated our proposed modality fusion segmentation network over two LiDAR sequential point cloud datasets that were collected under the autonomous diving scenario. Both datasets have been widely used in the literature of point cloud segmentation.
\begin{itemize}
    \item \textit{SemanticKITTI:} The SemanticKITTI dataset \cite{behley2019semantickitti} is a newly published large scale semantic segmentation LiDAR point cloud dataset by labeling point-wise annotation for raw data of odometry task in KITTI benchmark \cite{geiger2012kitti}. It consists of 11 sequences for training (sequences 00-10) and 11 sequences for testing (sequences 11-21). The whole dataset contains over 43,000 scans and each scan has more than 100k points in average. Labels over 19 classes for each points on the training set are provided and users should upload prediction on the testing set to official website\footnote{http://semantic-kitti.org/index.html} for a fair evaluation.
    \item \textit{KITTI:} Wu et al.\cite{wu2018squeezeseg} introduced a semantic segmentation dataset exported from LiDAR data in KITTI detection benchmark \cite{geiger2012kitti}. This dataset was collected by using the same LiDAR sensor as SemanticKITTI but in different scenes. It is split into a training set with 8,057 spherical projected 2D images and a testing set with 2,791 projected images. Semantic labels over \textit{car, pedestrian} and \textit{bicycle} are provided.
\end{itemize}

\textbf{Implementation details:} The resolution of the projected images is set as $(H=64,W=2048)$ for SemanticKITTI and $(H=64,W=512)$ for KITTI, which are the same as previous methods \cite{wu2018squeezeseg,wu2019squeezesegv2,milioto2019rangenet++, biasutti2019lunet} for fair comparisons.
We follow the same way to split dataset as previous methods. Further, for a faster experiments, we uniformly sample 1/4 scans in training set as \textit{sub-SemanticKITTI} for training and remain whole validation set for testing in experiments of ablation studies and discussion. We train FPS-Net for 200 epochs with batch size of 16 using Adam optimizer with initial learning rate $lr=0.001$. For data augmentation on each scan, We randomly rotate each scan within $[-1^\circ,1^\circ]$ and flip horizontally. $\lambda$ is set to be 1. 

\textbf{Evaluation Metrics:} Performances are evaluated via per-class IoU and mean IoU (mIoU). For class $c$, the intersection over union (IoU) is defined as the intersection between class prediction $\mathcal{P}_c$ and ground truth $\mathcal{G}_c$ divided by their union part
\begin{equation}
    {IoU}_{c}=\frac{\mathcal{P}_c \cap \mathcal{G}_c}{\mathcal{P}_c \cup \mathcal{G}_c}
\end{equation}
And the mIoU is the average of IoU over all classes
\begin{equation}
    mIoU=\frac{1}{C}\sum_{c=1}^{C}{{IoU}_c}
\end{equation}

We use the Frame Per Scan (FPS) to evaluate the efficiency of different cloud segmentation methods. For fair comparison, all of inference FPS results including our model and compared models are evaluated with a GeForce RTX 2080 TI Graphics Card.

\subsection{Experimental Results}\label{Sec.benchmarks_results}
\subsubsection{SemanticKITTI}

We compare our FPS-Net with 
state-of-the-art projection-based approaches on the test set of SemanticKITTI benchmark, and Table \ref{tab.SOTA-comparison-semantickitti} shows experimental results. As we can see, classical projection-based methods including SqueezeSeg~\cite{wu2018squeezeseg} and SqueezeSegV2~\cite{wu2019squeezesegv2} achieve very efficient computation but both of them sacrificed accuracy greatly. The recently proposed RangeNet++~\cite{milioto2019rangenet++}, PolarNet~\cite{zhang2020polarnet} and SqueezeSegV3~\cite{xu2020squeezesegv3} achieved higher accuracy but their processing speed dropped instead. As a comparison, Our FPS-Net achieves the best segmentation accuracy in mIoU, and its efficiency is also comparable with the state-of-the-art models.

Specifically, FPS-Net outperforms PolarNet \cite{zhang2020polarnet} and SequeezeSegV3 \cite{xu2020squeezesegv3} in both mIoU (+2.8\% and +1.2\%) and computational efficiency (+5.2 FPS and +15.0 FPS). 
By learning modality-specific features from projected images, FPS-Net can capture detailed information from respective modality. The hierarchical leaning strategy in both MRF-RDB and RCB helps to keep features of small objects with fewer points. As a results, FPS-Net improves the segmentation accuracy by large margins for object classes \textit{bicycle, motorcycle, person and bicyclist}. It also achieves the best accuracy for other classes such as \textit{traffic-sign and truck}.

\renewcommand\arraystretch{1.4}
\begin{table}[t]
  \caption{Comparison of FPS-Net with state-of-the-art projection-based methods on the test set of SemanticKITTI: Both FPS and mIoU are reported for each compared method. Note FPS values were re-produced with the same configuration for fair comparison and they may not be consistent with the original papers.}
  \centering
  \scriptsize
  \setlength{\tabcolsep}{0.6mm}{
  \begin{tabular}{r|cc|ccccccccccccccccccc}
    \hline
    Model & \rotatebox{90}{\textbf{FPS}} & \rotatebox{90}{\textbf{mIoU}} & \rotatebox{90}{road} & \rotatebox{90}{sidewalk} & \rotatebox{90}{parking} & \rotatebox{90}{other-ground} & \rotatebox{90}{building} & \rotatebox{90}{car} & \rotatebox{90}{truck} & \rotatebox{90}{bicycle} & \rotatebox{90}{motorcycle} & \rotatebox{90}{other-vehicle} & \rotatebox{90}{vegetation} & \rotatebox{90}{trunk} & \rotatebox{90}{terrain} & \rotatebox{90}{person} & \rotatebox{90}{bicyclist} & \rotatebox{90}{motorcyclist} & \rotatebox{90}{fence} & \rotatebox{90}{pole} & \rotatebox{90}{traffic-sign}\\
    \hline
    SqueezeSeg\cite{wu2018squeezeseg} & 72.6 & 29.5 & 85.4 & 54.3 & 26.9 & 4.5 & 57.4 & 68.8 & 3.3 & 16.0 & 4.1 & 3.6 & 60.0 & 24.3 & 53.7 & 12.9 & 13.1 & 0.9 & 29.0 & 17.5 & 24.5\\
    SqueezeSegV2\cite{wu2019squeezesegv2} & 62.8 & 39.7 & 88.6 & 67.6 & 45.8 & 17.7 & 73.7 & 81.8 & 13.4 & 18.5 & 17.9 & 14.0 & 71.8 & 35.8 & 60.2 & 20.1 & 25.1 & 3.9 & 41.1 & 20.2 & 36.3\\
    RangeNet++\cite{milioto2019rangenet++} & 14.9 & 52.2 & \textbf{91.8} & \textbf{75.2} & \textbf{65.0} & \textbf{27.8} & 87.4 & 91.4 & 25.7 & 25.7 & 34.4 & 23.0 & 80.5 & 55.1 & 64.6 & 38.3 & 38.8 & 4.8 & 58.6 & 47.9 & 55.9 \\
    PolarNet\cite{zhang2020polarnet} & 15.6 & 54.3 & 90.8 & 74.4 & 61.7 & 21.7 & \textbf{90.0} & \textbf{93.8} & 22.9 & 40.3 & 30.1 & 28.5 & \textbf{84.0} & \textbf{65.5} & \textbf{67.8} & 43.2 & 40.2 & 5.6 & \textbf{67.8} & \textbf{51.8} & 57.5 \\
    SqueezeSegv3\cite{xu2020squeezesegv3} & 5.8 & 55.9 & 91.7 & 74.8 & 63.4 & 26.4 & 89.0 & 92.5 & 29.6 &38.7 & 36.5 & \textbf{33.0} & 82.0 & 58.7 & 65.4 & 45.6 & 46.2 & \textbf{20.1} & 59.4& 49.6 & 58.9\\
    \hline
    \textbf{FPS-Net (ours)} & 20.8 & \textbf{57.1} &  91.1 & 74.6 & 61.9 & 26.0 & 87.4 & 91.1 & \textbf{37.1} & \textbf{48.6} & \textbf{37.8} & 30.0 & 80.9 & 61.2 & 65.0 & \textbf{60.5} & \textbf{57.8} & 7.5 & 57.4 & 49.9 & \textbf{59.2} \\
    \hline
  \end{tabular}
  }
  \label{tab.SOTA-comparison-semantickitti}
\end{table}

\renewcommand\arraystretch{1.2}
\begin{table}[t]
  \caption{Comparison of FPS-Net with state-of-the-art projection-based methods on the validation set of KITTI}
  \centering
  \begin{tabular}{r|c|c|ccc}
    \hline
    Model & size & mIoU & car & cyclist & pedestrian \\
    \hline
    SqueezeSeg\cite{wu2018squeezeseg} & $64\times512$ & 37.2 & 64.6 & 21.8 & 25.1 \\
    PointSeg \cite{wang2018pointseg} & $64\times512$ & 39.8 & 67.4 & 32.7 & 19.2\\
    SqueezeSegv2\cite{wu2019squeezesegv2} & $64\times512$ & 44.9 & 73.2 & 27.8 & 33.6 \\
    LuNet\cite{biasutti2019lunet} & $64\times512$ & 55.4 & 72.7 & 46.9 & 46.5 \\
    RangeNet++\cite{milioto2019rangenet++} & $64\times512$ & 57.3 & 97.0 & 44.0 & 30.8 \\
    \hline
   \textbf{FPS-Net (Ours)} & $64\times512$ & \textbf{69.9} & \textbf{98.7} & \textbf{61.5} & \textbf{49.4} \\
    \hline
  \end{tabular}
  \label{tab.SOTA-comparison-kitti}
\end{table}

\begin{figure}
  \centering
  \begin{subfigure}[b]{0.48\textwidth}
  \includegraphics[width=\textwidth]{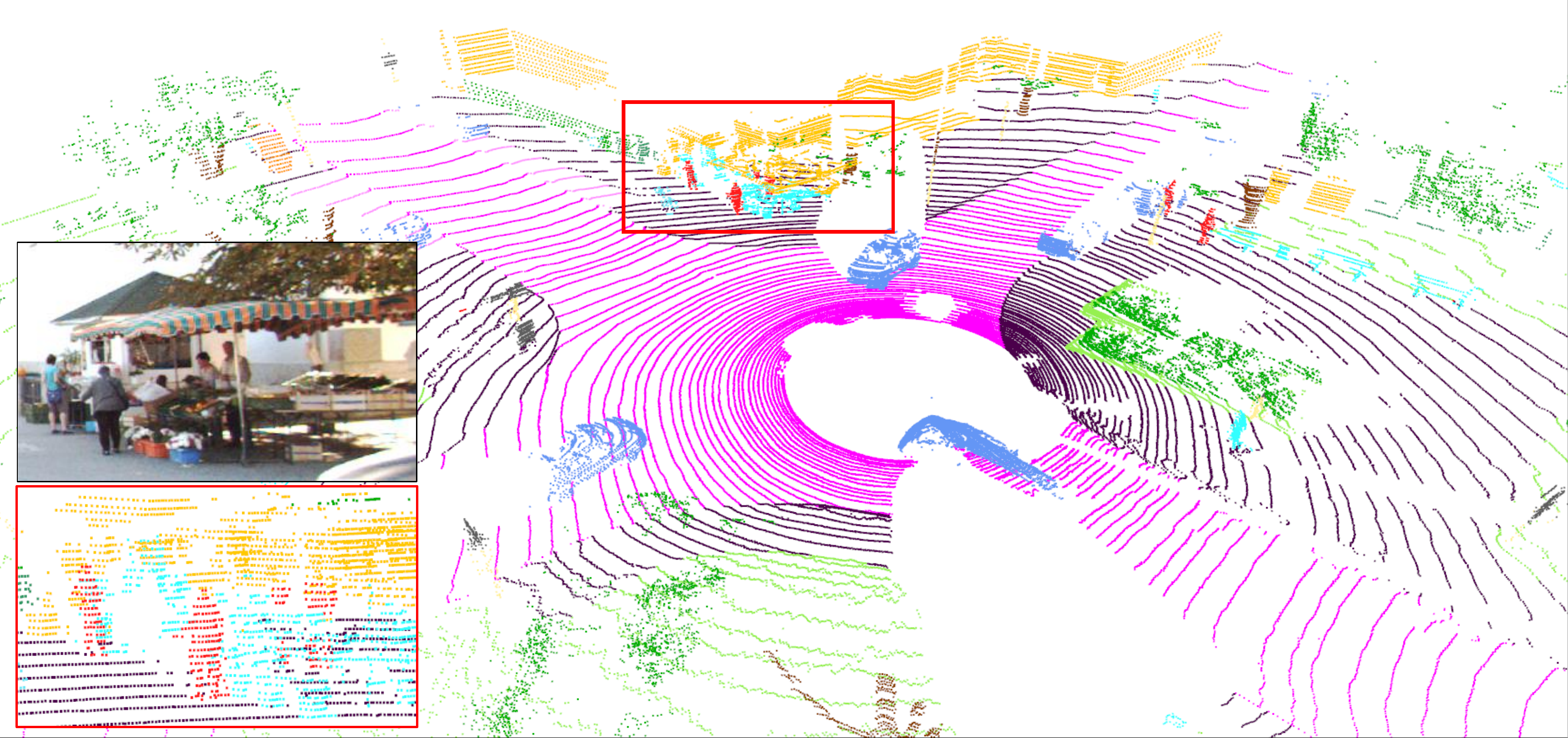}
  \caption{Prediction \#1}
  \end{subfigure}
  \begin{subfigure}[b]{0.48\textwidth}
  \includegraphics[width=\textwidth]{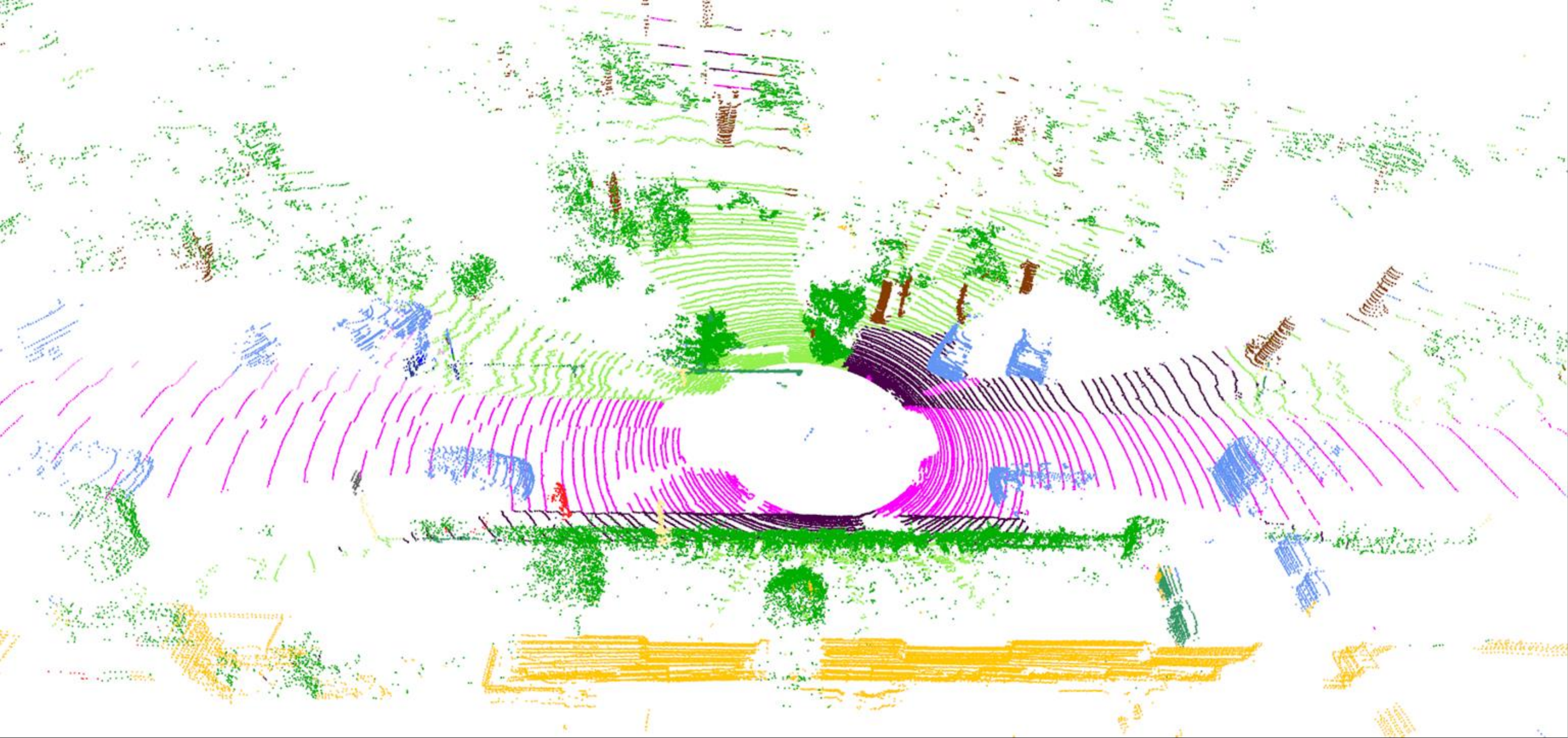}
  \caption{Prediction \#2}
  \end{subfigure}
 \begin{subfigure}[b]{0.48\textwidth}
  \includegraphics[width=\textwidth]{fig5_51_gt.pdf}
  \caption{Ground Truth \#1}
  \end{subfigure}
  \begin{subfigure}[b]{0.48\textwidth}
  \includegraphics[width=\textwidth]{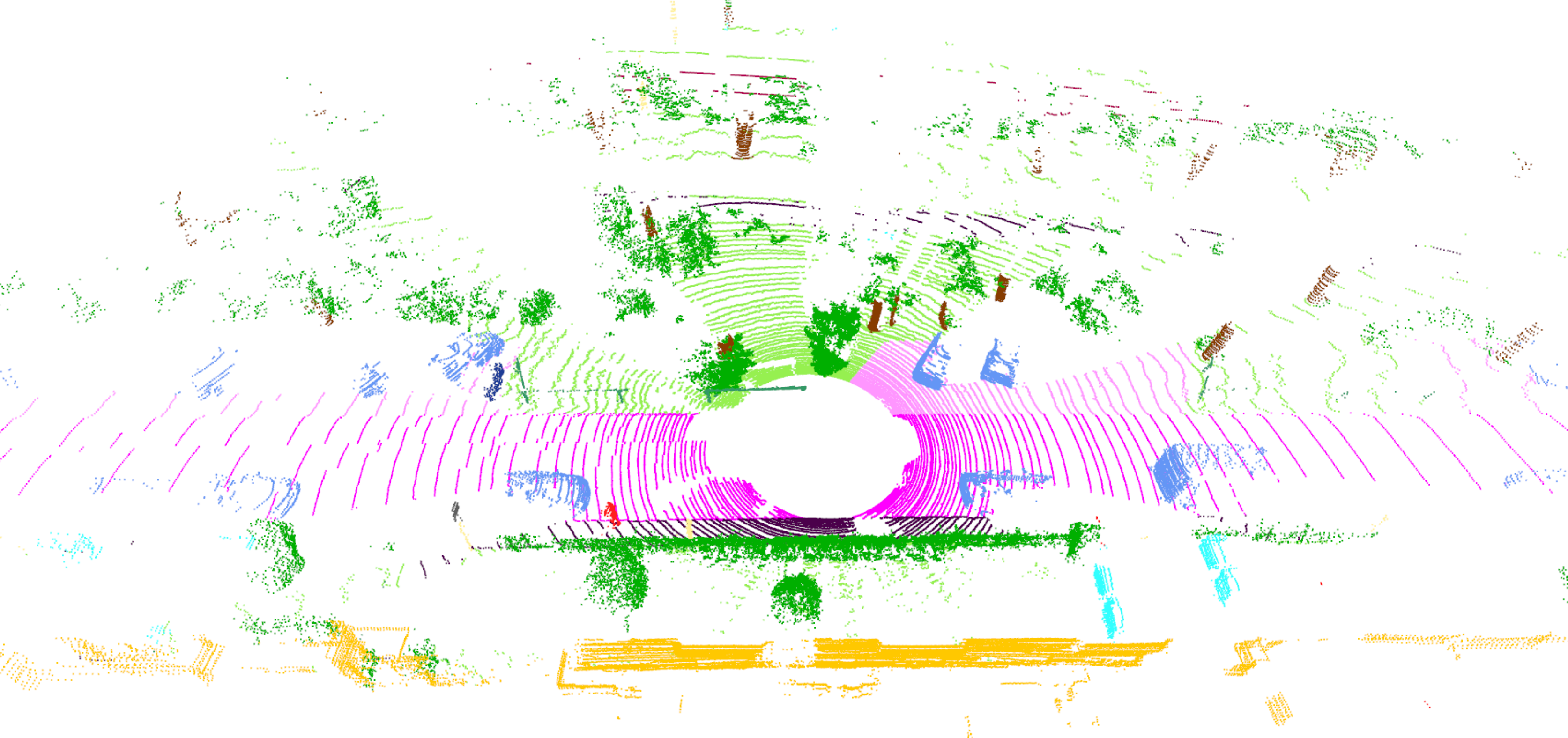}
  \caption{Ground Truth \#2}
  \end{subfigure}
  \begin{subfigure}[b]{0.24\textwidth}
  \includegraphics[width=\textwidth]{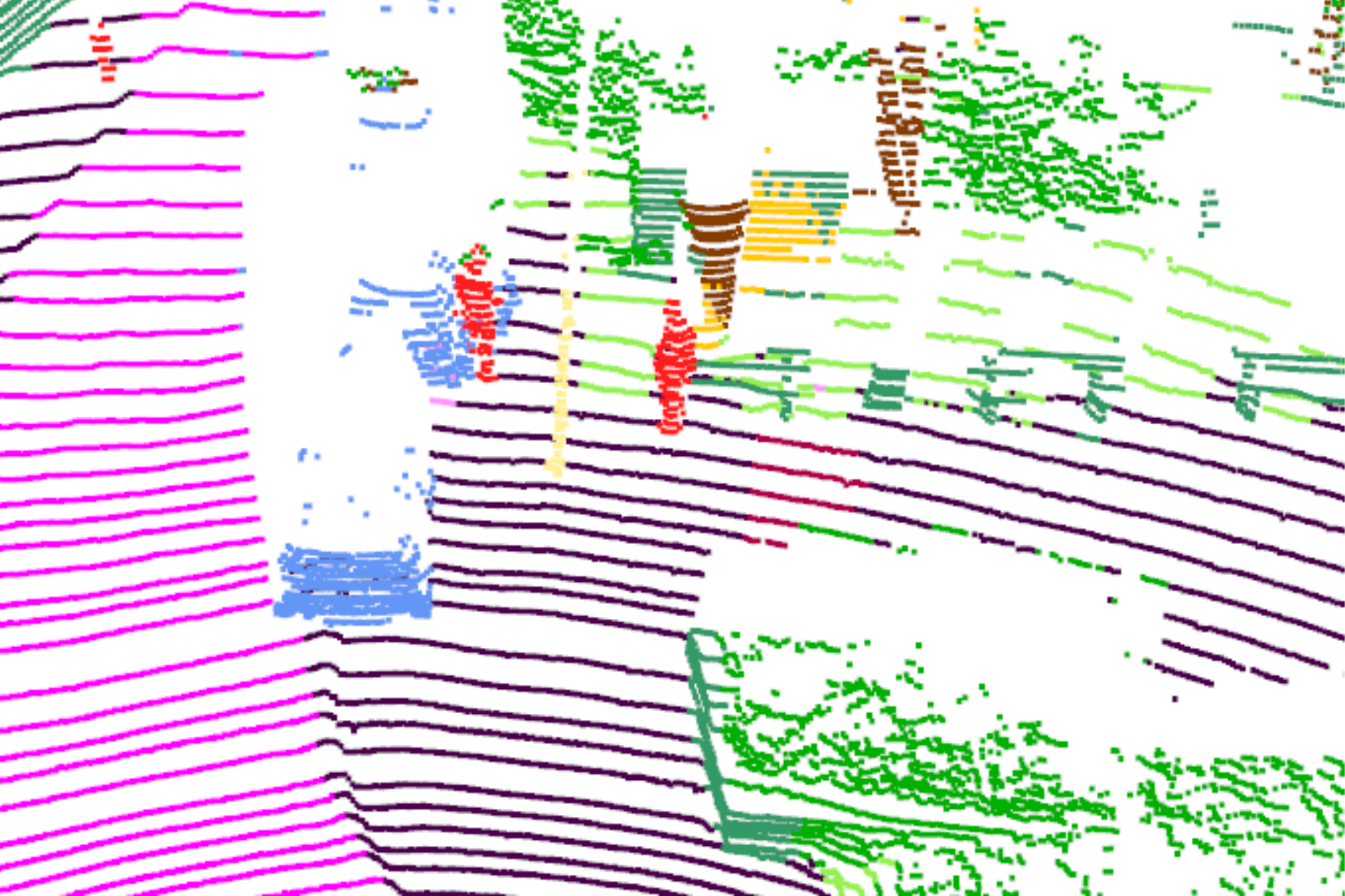}
  \caption{Prediction \#3}
  \end{subfigure}
  \begin{subfigure}[b]{0.24\textwidth}
  \includegraphics[width=\textwidth]{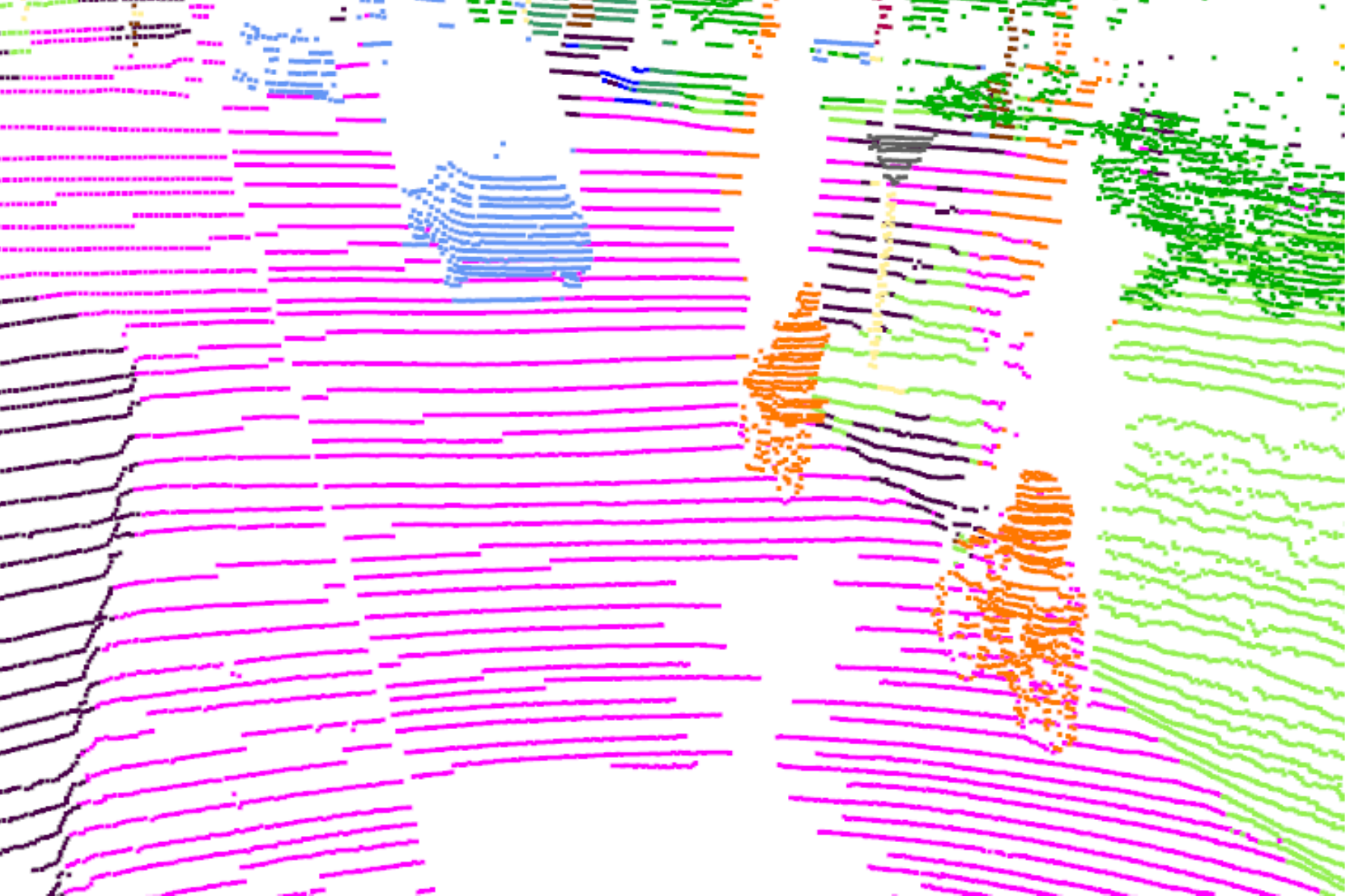}
  \caption{Prediction \#4}
  \end{subfigure}
  \begin{subfigure}[b]{0.24\textwidth}
  \includegraphics[width=\textwidth]{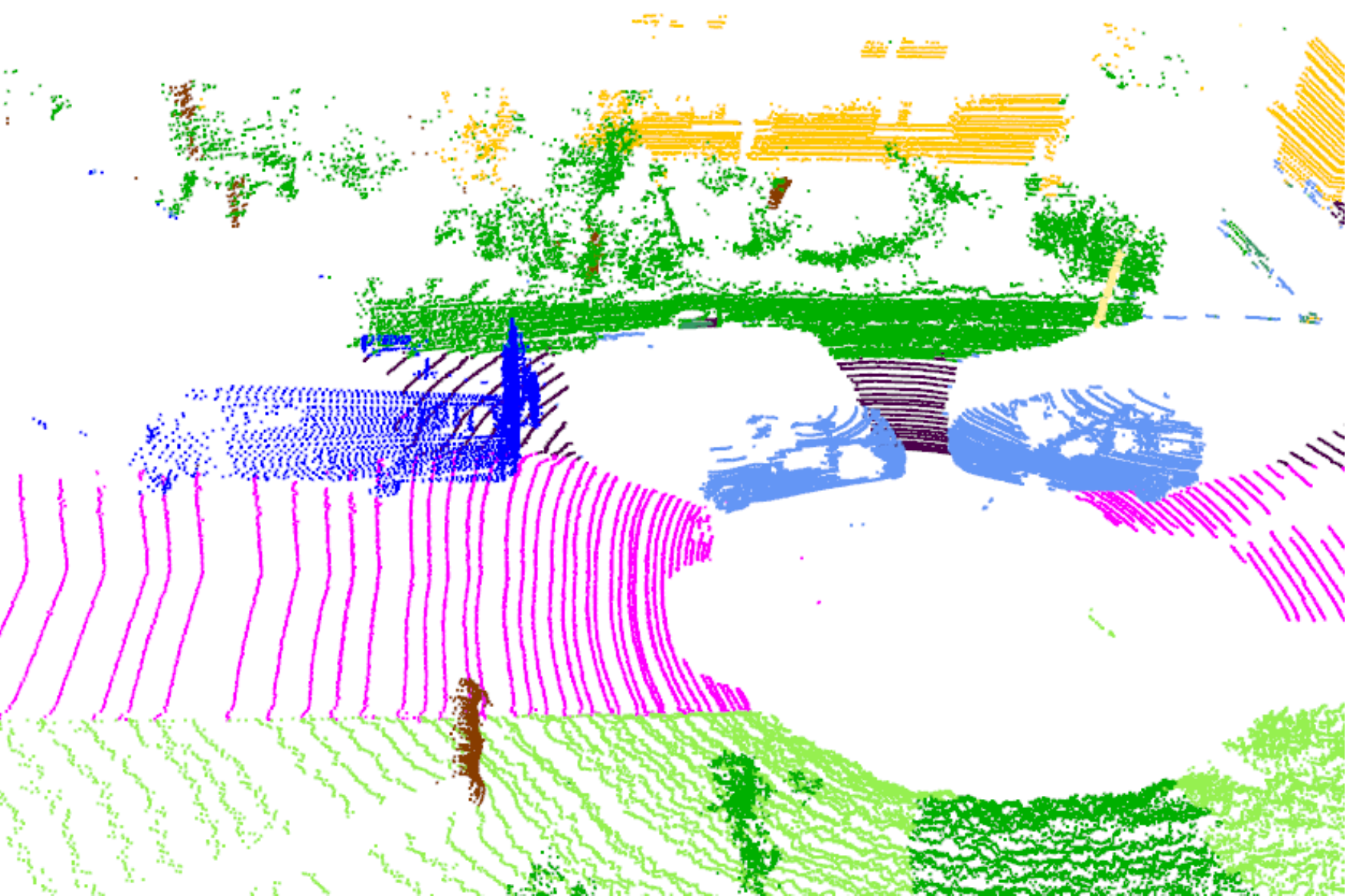}
  \caption{Prediction\# 5}
  \end{subfigure}
  \begin{subfigure}[b]{0.24\textwidth}
  \includegraphics[width=\textwidth]{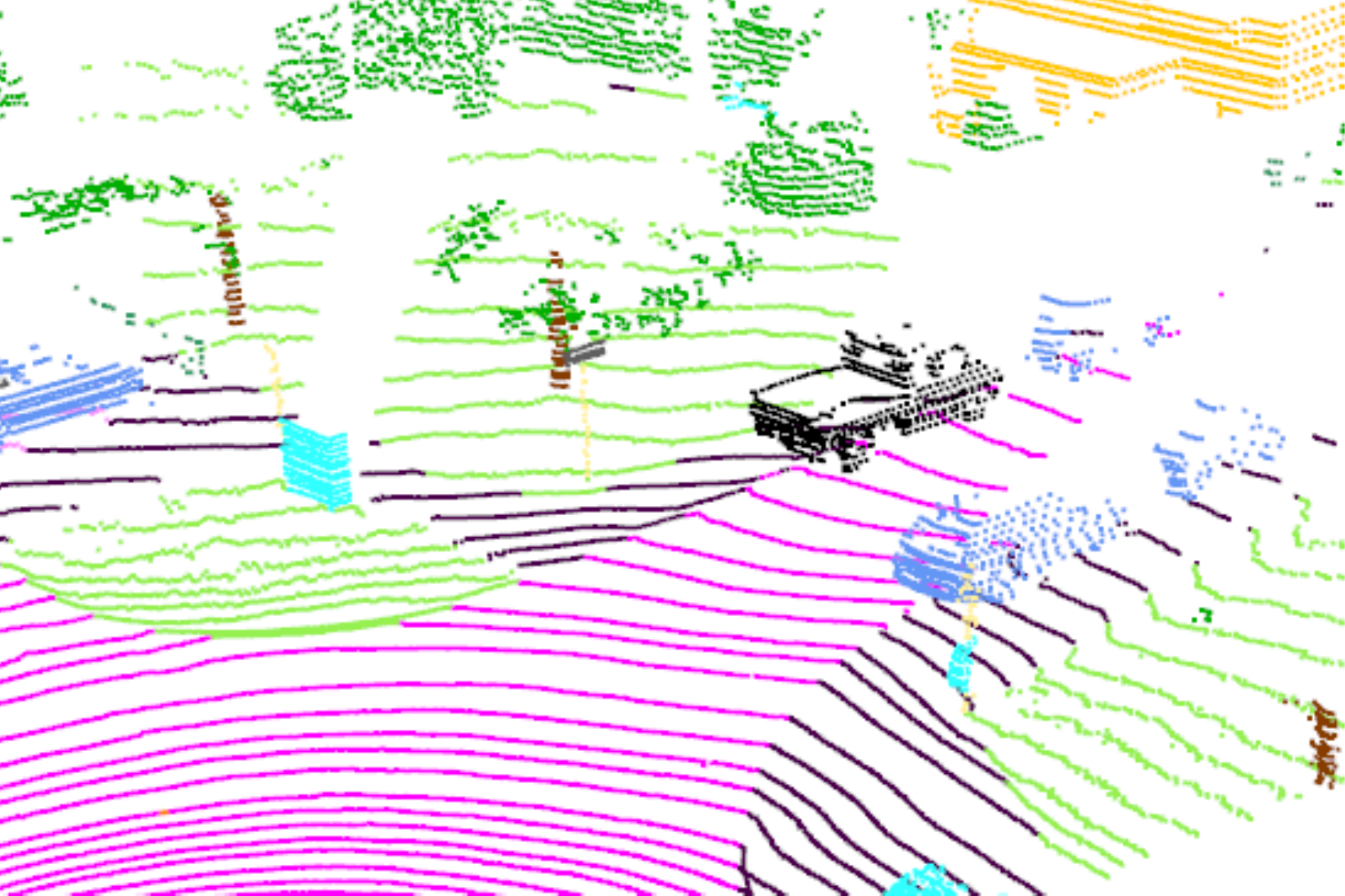}
  \caption{Prediction \#6}
  \end{subfigure}
  \begin{subfigure}[b]{0.24\textwidth}
  \includegraphics[width=\textwidth]{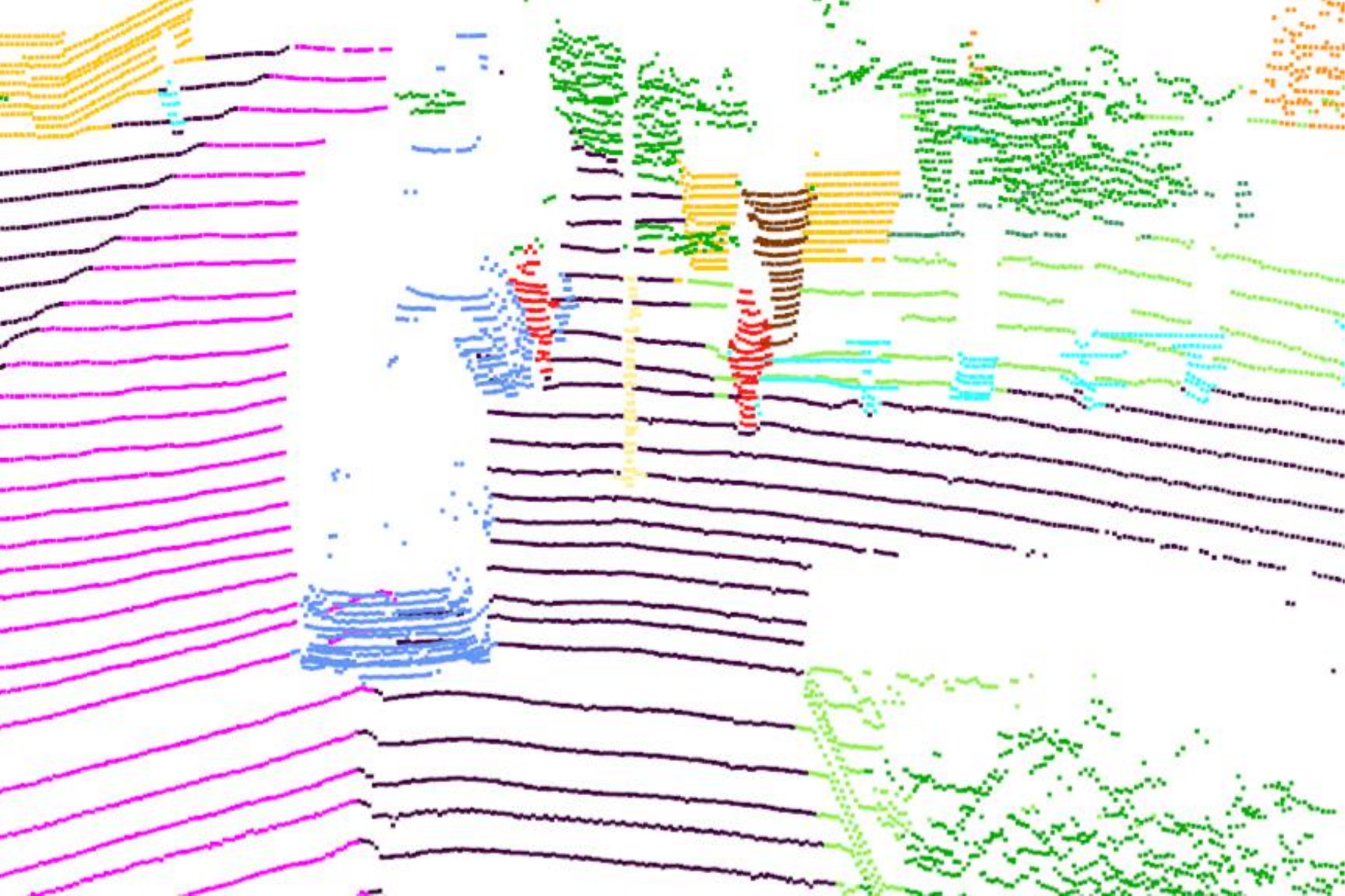}
  \caption{Ground Truth \#3}
  \end{subfigure}
  \begin{subfigure}[b]{0.24\textwidth}
  \includegraphics[width=\textwidth]{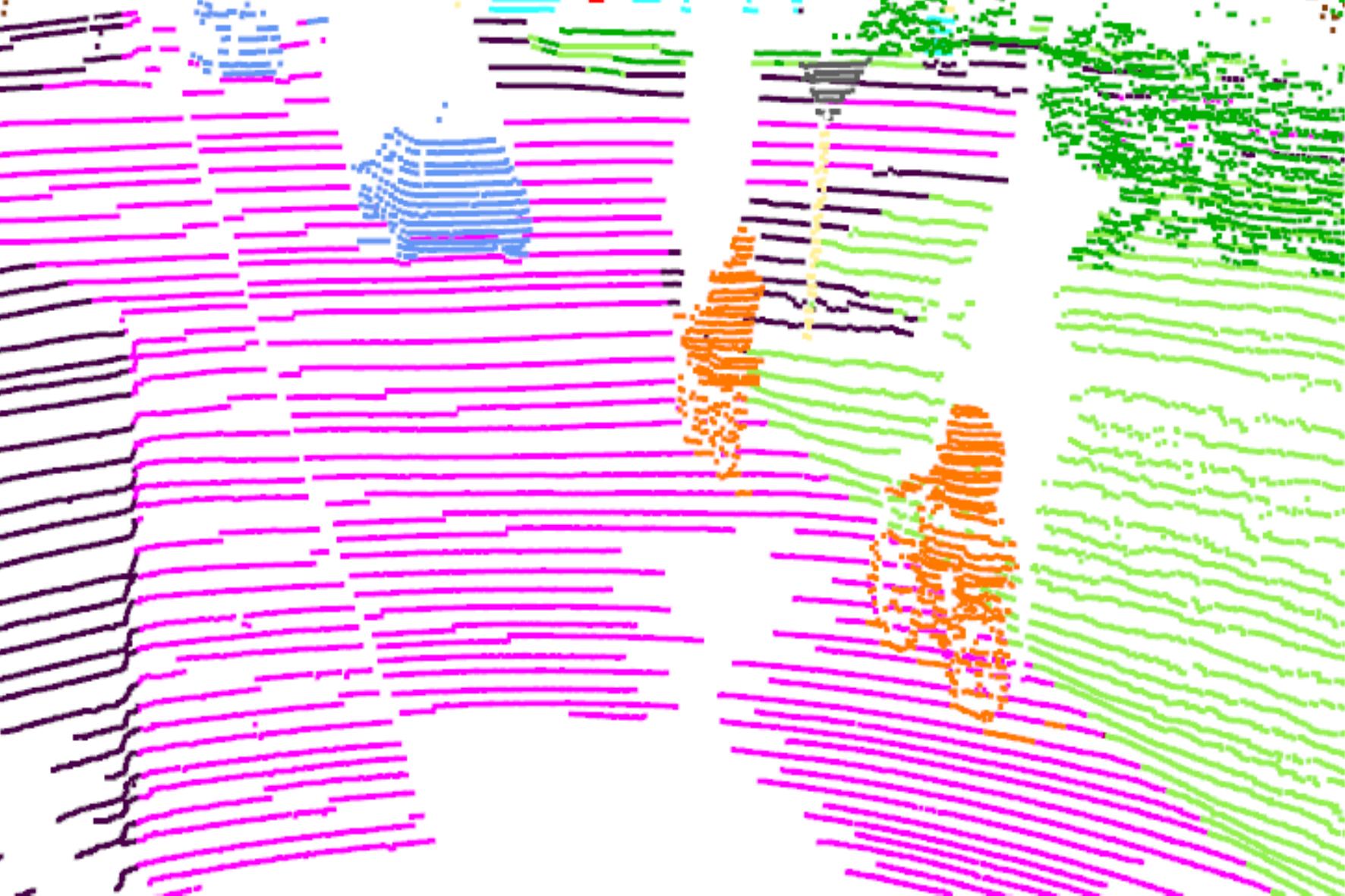}
  \caption{Ground Truth \#4}
  \end{subfigure}
  \begin{subfigure}[b]{0.24\textwidth}
  \includegraphics[width=\textwidth]{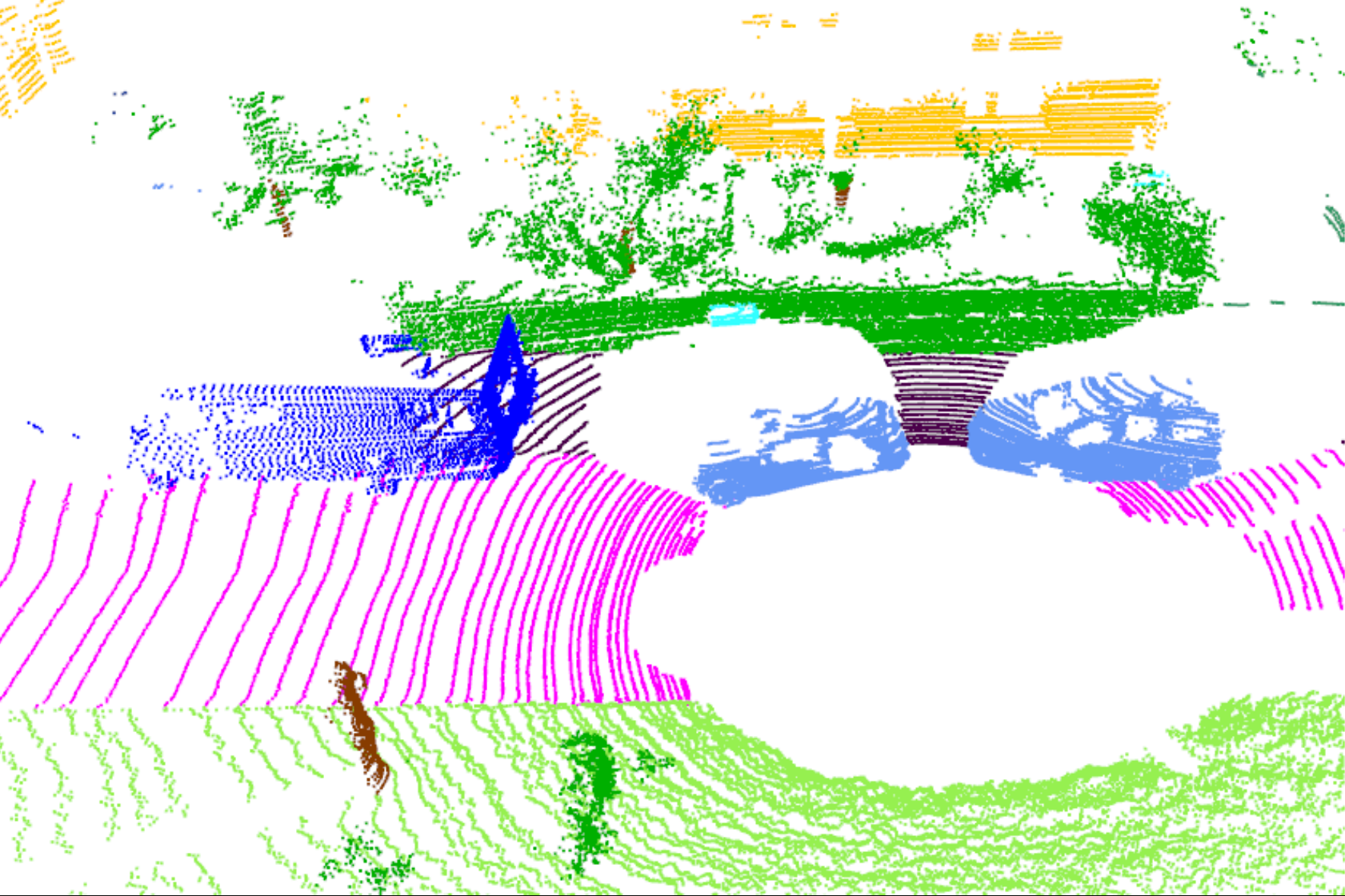}
  \caption{Ground Truth \#5}
  \end{subfigure}
  \begin{subfigure}[b]{0.24\textwidth}
  \includegraphics[width=\textwidth]{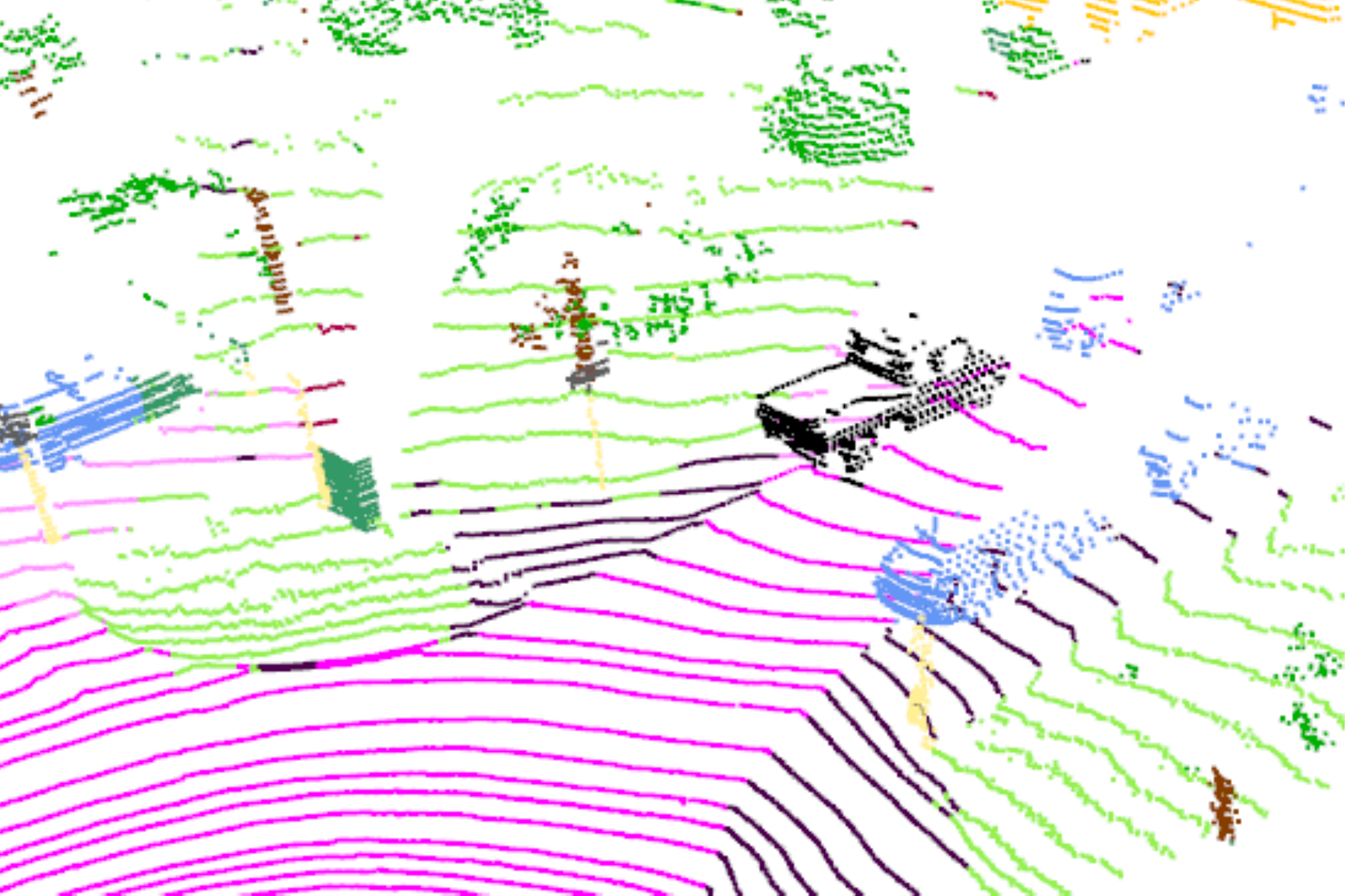}
  \caption{Ground Truth \#6}
  \end{subfigure}
  \begin{subfigure}[b]{0.9\textwidth}
  \includegraphics[width=\textwidth]{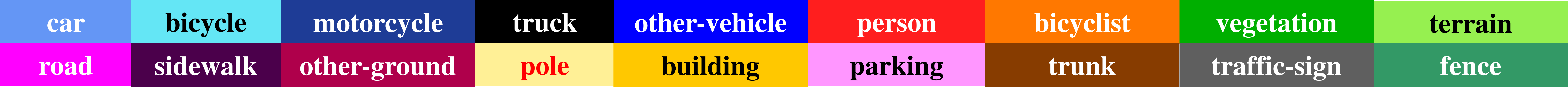}
  \caption{Legend}
  \end{subfigure}
  \caption{Illustration of semantic segmentation of SemanticKITTI point cloud by FPS-Net. (a) and (b) show predictions of two point cloud scans; (e), (f), (g), (h) are detailed segmentation of foreground classes; (c), (d), (i), (j), (k), (h) show the corresponding ground truth. FPS-Net can correctly classify most points which is well aligned with the results in Table \ref{tab.SOTA-comparison-semantickitti}. Segmentation is more prone to errors around the class-transition regions, as highlighted in red bounding boxes of (a) and (c) with their close-up views shown at the lower-left corner (including the corresponding visual images for easy interpretation).}
  \label{figure.vis_semantickitti}
\end{figure}

Fig. \ref{figure.vis_semantickitti} illustrates FPS-Net with several point cloud samples. Specifically, Figs. \ref{figure.vis_semantickitti}(a) and (b) show predictions of two point cloud scans in SemanticKITTI and Figs. \ref{figure.vis_semantickitti}(c) and (d) show the corresponding ground truth (all colored by classes as legend (m)), respectively. It can be seen that FPS-Net can correctly classify most points especially classes of \textit{road, sidewalk, building, car, and vegetation}. On the other hand, it tends to produce segmentation errors around the class-transition regions due to a mixture in classes and lower point density there which makes point cloud segmentation a more challenging task. A typical example is highlighted in red bounding boxes of (a) and (c) with their close-up views shown at the lower-left corner (including the corresponding visual images for easy interpretation). We can observe that points around this region belong to different classes and many of them are classified as \textit{person} incorrectly. In addition, Figs. \ref{figure.vis_semantickitti}(e), (f), (g), (h) show four detailed segmentation of classes \textit{person, bicyclist, car, other-vehicle} and \textit{truck}. The segmentation is well aligned with the results in Table \ref{tab.SOTA-comparison-semantickitti}.

\subsubsection{KITTI} 

We also compare FPS-Net with state-of-the-art methods over the validation set of KITTI semantic segmentation dataset, and Table \ref{tab.SOTA-comparison-kitti} shows experimental results. It is obvious that FPS-Net outperforms other models by large margins in the metric of mIoU (+12.6\%). For specific classes, our model achieves large improvement on \textit{bicycle} (+14.6\%) as compared with the state-of-the-art and smaller increases in \textit{pedestrian} (+2.9\%) and \textit{car} (+1.7\%). The experimental results are well aligned with that over the SemanticKITTI as our hierarchical learning learns useful features for classes with fewer points and improves more over those classes. They also show that FPS-Net is robust and can produce superior segmentation across different datasets consistently.

\subsection{Ablation Studies}\label{Sec.ablation_study}
We performed extensive ablation studies to evaluate individual FPS-Net components separately. In our ablation study, we uniformly sampled 1/4 data in the SemanticKITTI training set (named as '\textit{sub-SemanticKITTI}') in training (for faster training) and the whole validation set for inference.

\renewcommand\arraystretch{1.3}
\begin{table}[ht] 
  \caption{Ablation study of FPS-Net over the SemanticKITTI validation set: \textit{Stacked learning} means stacking multi-modal data as a single input while \textit{separate learning} refers to our method in FPS-Net. \textit{RDB}, \textit{MRF-RDB} and \textit{RCB} are building blocks as introduced in Section \ref{Sec.domain_fusion_learn}.}
  \centering
  \begin{tabular}{ccccc}
    \hline
    Input & Encoder & Decoder & Fusion & mIoU \\
    \hline
    \multirow{4}*{stacked learning} & RCB & RCB & None & 46.5\\
    ~ & RDB & RDB & None &  49.3\\
    ~ & RDB & RCB & None &  50.0\\
    ~ & MRF-RDB & RCB & None &  52.3\\
    \hline
    separate learning & MRF-RDB & RCB & Fused & 54.9\\
    \hline
 \end{tabular}
 \label{tab.components}
\end{table}

\renewcommand\arraystretch{1.4}
\begin{table}[!h]
  \caption{Ablation study of different modality combinations (on SemanticKITTI validation set with image size=$1024\times64$): '-' means only one modality image is used, 'stacked' means stacking multi-modal data as one input, and 'fused' refers to our separate learning and fusing method.}
  \centering
  \scriptsize
  \setlength{\tabcolsep}{0.8mm}{
  \begin{tabular}{ccc|c|c|ccccccccccccccccccc}
    \hline
    xyz & intensity & depth & type & mIoU &
    \rotatebox{90}{car} & \rotatebox{90}{bicycle} & \rotatebox{90}{motorcycle} & \rotatebox{90}{truck}  & \rotatebox{90}{other-vehicle} & \rotatebox{90}{person} & \rotatebox{90}{bicyclist} & \rotatebox{90}{motorcyclist} & \rotatebox{90}{road} & \rotatebox{90}{parking} & \rotatebox{90}{sidewalk} & \rotatebox{90}{other-ground} & \rotatebox{90}{building} & \rotatebox{90}{fence} & \rotatebox{90}{vegetation} & \rotatebox{90}{trunk} & \rotatebox{90}{terrain} & \rotatebox{90}{pole} & \rotatebox{90}{traffic-sign}\\
    \hline
    \checkmark & & & - & 47.3 & 86.9 & 18.0 & 24.8 & 48.4 & 26.9 & 42.4 & 58.4 & 0.0 & 90.8 & 39.9 & 73.3 & 0.3 & 78.3 & 40.5 & 78.9 & 48.6 & 64.2 & 46.5 & 32.0\\
     & \checkmark & & - & 39.5 & 78.9 & 21.9 & 11.9 & 30.0 & 25.4 & 28.1 & 43.3 & 0.0 & 86.9 & 27.7 & 67.7 & 0.1 & 71.0 & 33.9 & 75.3 & 34.3 & 62.9 & 24.6 & 27.5\\
     & & \checkmark & - & 49.3 & 88.1 & 18.3 & 22.8 & 69.1 & 32.4 & 45.7 & 52.4 & 0.0 & 91.0 & 35.1 & 73.9 & 1.0 & 80.5 & 44.5 & 80.0 & 52.8 & 67.0 & 46.6 & 35.5\\
    \hline
    \checkmark &  & \checkmark & stacked & 46.5 & 87.2 & 11.2 & 21.1 & 58.5 & 28.8 & 39.6 & 51.2 & 0.0 & 90.3 & 36.4 & 72.3 & 0.0 & 78.1 & 38.5 & 77.5 & 50.2 & 64.1 & 45.7 & 32.5\\
    \checkmark & \checkmark & & stacked & 49.3 & 87.3 & 33.2 & 21.5 & 63.3 & 27.1 & 45.3 & 54.7 & 0.0 & 90.8 & 36.5 & 75.6 & 0.1 & 79.5 & 38.1 & 79.9 & 52.2 & 69.1 & 44.4 & 36.8\\
    & \checkmark & \checkmark & stacked & 50.6 & 87.6 & 31.7 & 26.6 & 66.8 & 28.0 & 46.9 & 60.8 & 0.0 & 90.1 & 34.5 & 74.3 & 0.2 & 81.1 & 45.1 & 80.8 & 52.7 & 67.0 & 48.3 & 39.6\\
    \checkmark & \checkmark & \checkmark & stacked & 50.2 & 88.9 & 37.6 & 28.6 & 48.8 & 31.5 & 46.7 & 58.9 & 0.0 & 91.4 & 32.4 & 75.6 & 0.8 & 80.6 & 45.2 & 80.3 & 52.0 & 65.3 & 49.5 & 39.4\\
    \hline
    \checkmark &  & \checkmark & fused & 49.8 & 88.2 & 7.2 & 20.1 & 77.5 & 35.6 & 44.5 & 59.5 & 0.0 & 90.6 & 38.7 & 74.7 & 0.2 & 80.5 & 45.4 & 80.7 & 52.7 & 67.7 & 48.9 & 32.5\\
    \checkmark & \checkmark & & fused & 51.2 & 88.7 & 38.3 & 26.5 & 55.0 & 34.4 & 48.4 & 58.4 & 0.0 & 91.5 & 31.3 & 76.8 & 0.1 & 82.1 & 47.8 & 81.7 & 52.3 & 68.5 & 50.9 & 39.0\\
    & \checkmark & \checkmark & fused & 50.7 & 88.0 & 39.1 & 25.3 & 61.6 & 24.2 & 46.8 & 59.9 & 0.0 & 91.3 & 34.7 & 76.1 & 0.1 & 80.9 & 44.2 & 80.9 & 53.0 & 64.7 & 48.7 & 42.8\\
    \checkmark & \checkmark & \checkmark & fused & 52.0 & 88.6 & 38.3 & 26.4 & 58.1 & 31.6 & 55.4 & 60.9 & 0.1 & 91.9 & 34.8 & 76.5 & 0.1 & 81.6 & 45.1 & 82.5 & 54.2 & 68.8 & 50.7 & 41.6\\
    \hline
  \end{tabular}}
  \label{tab.Ablation-modality}
\end{table}
 
We first examine how different FPS-Net modules contribute to the overall segmentation performance.
Firstly, we evaluated the two building blocks RCB and RDB without considering modality gaps (the same as in previous methods), where the five channel images are stacked as a single input as labelled by (\textit{stacked learning}) in Rows 2-4 in Table \ref{tab.components}. We can see that the network achieves the lowest segmentation accuracy when both encoder and decoder employ RCB. As a comparison, replacing RCB by RDB in the encoder produces the best accuracy, which means that the network performs much better by concatenating instead of adding for low- and high-level features in the encoder. In addition, Rows 2 and 3 show that using RDB in the decoder degrades the segmentation performance slightly as compared with using RCB in the decoder. This means that a simpler memory mechanism makes the model more discriminative in the ensuing classification step. Further, Rows 4 and 5 show that including a multi-receptive field module in the encoder (i.e. MRF-RDB) helps to improve the segmentation mIoU greatly by $2.3\%$. On top of that we use the separate learning and fusion approach aiming for tackling the modality gap problem, and the result shows that the segmentation performance increases another 2.6\% percentage (As the setup \textit{fused learning} in Table \ref{tab.components}). 

In addition, we studied the contribution of each modality in the fusion scheme. Table \ref{tab.Ablation-modality} shows experimental results, where Rows 2-4 show the segmentation while training models by using data of each of the three modalities, respectively. It can be seen that intensity performs much worse as compared with either depth or coordinate.
Additionally, depth performs better than coordinates, largely because the three coordinate channels (x, y, z) are closely correlated and CNNs cannot learn such correlation well. 
Rows 5-8 show the segmentation while stacking different modalities in a single image as input. 
We can see that intensity is complementary to both coordinate and depth while stacking coordinate and depth performs clearly worse than either modality alone. 
The last four rows show segmentation by FPS-Net. It can be seen that FPS-Net outperforms the stacking approach consistently under all modality combinations.
Additionally, fusing all three modalities achieves the best segmentation which further verifies the modality gap as well as the effectiveness of our proposed method.
Note stacking depth and intensity performs similarly as fusing them, indicating smaller gaps in between the two modalities.

\begin{figure}
  \begin{subfigure}[b]{0.45\textwidth}
  \includegraphics[width=\textwidth]{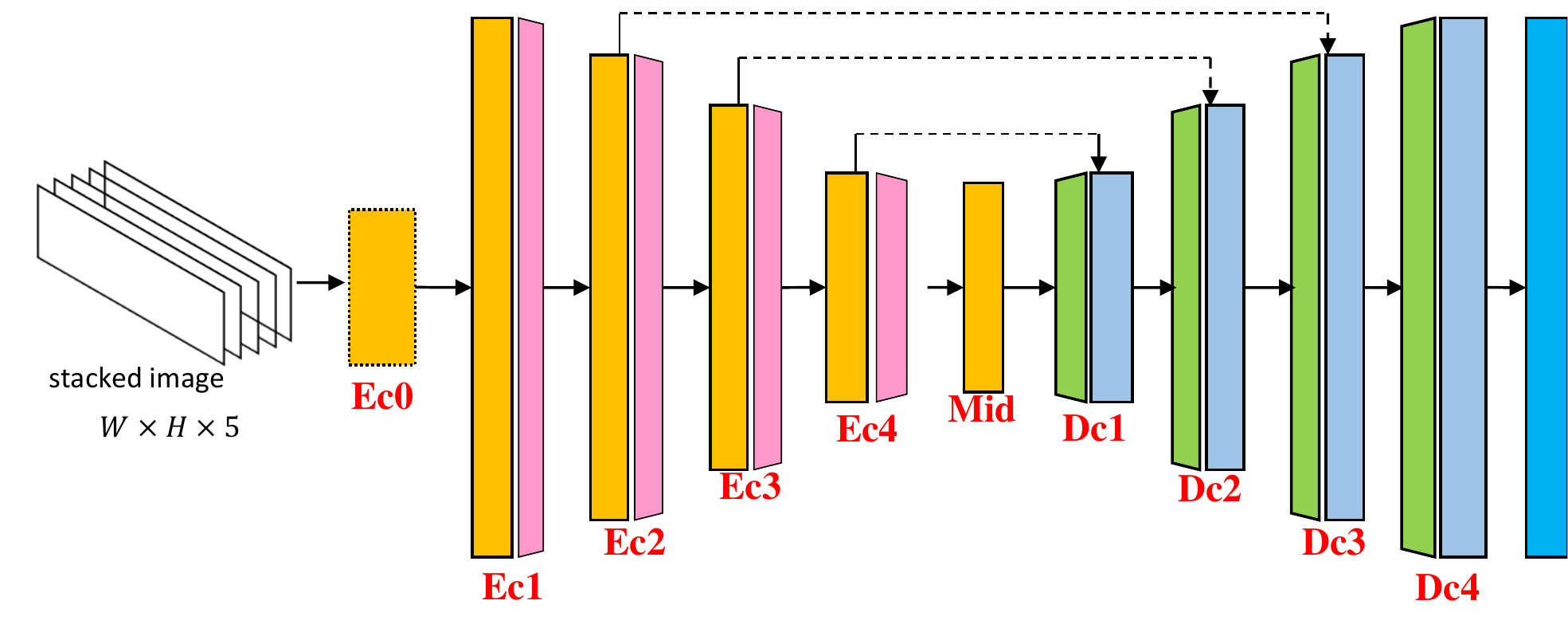}
  \caption{\textit{stacked} model}
  \end{subfigure}
  \begin{subfigure}[b]{0.45\textwidth}
  \includegraphics[width=\textwidth]{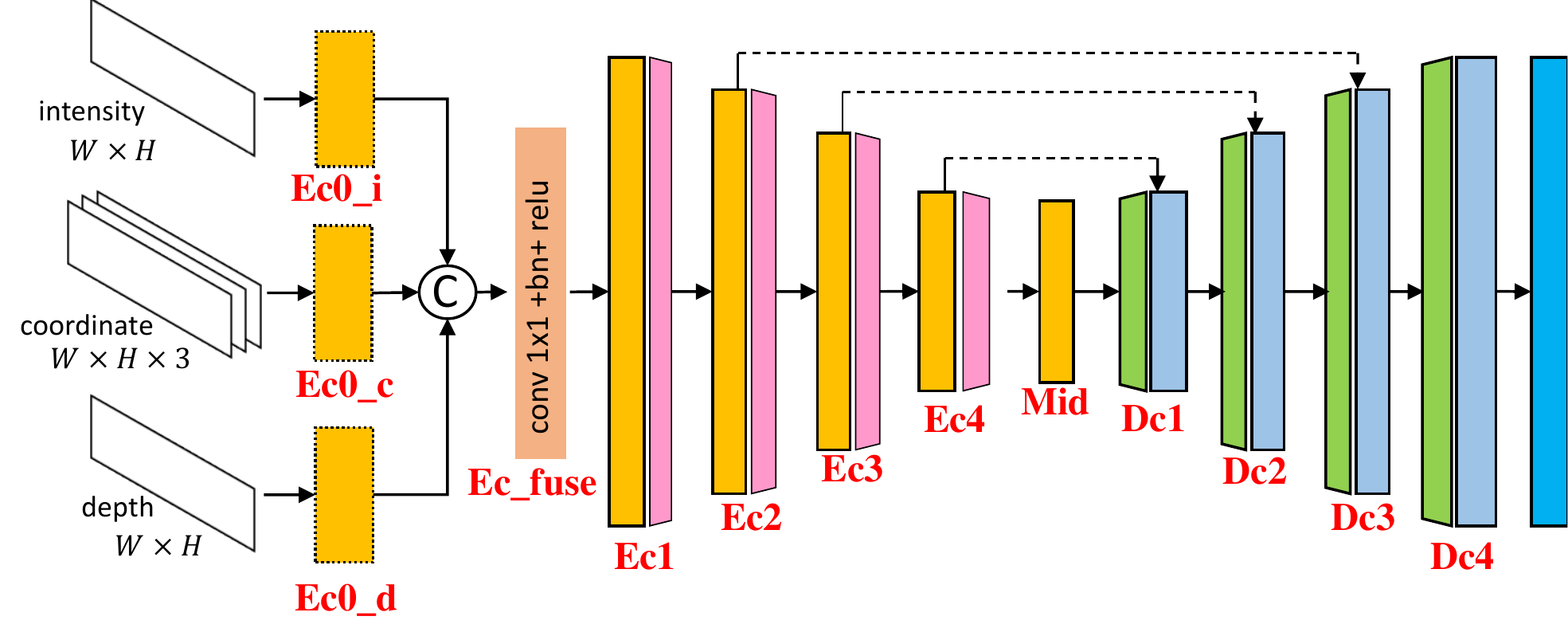}
  \caption{\textit{fusion} model}
  \end{subfigure}
  
  \begin{subfigure}[b]{\textwidth}
  \includegraphics[width=\textwidth]{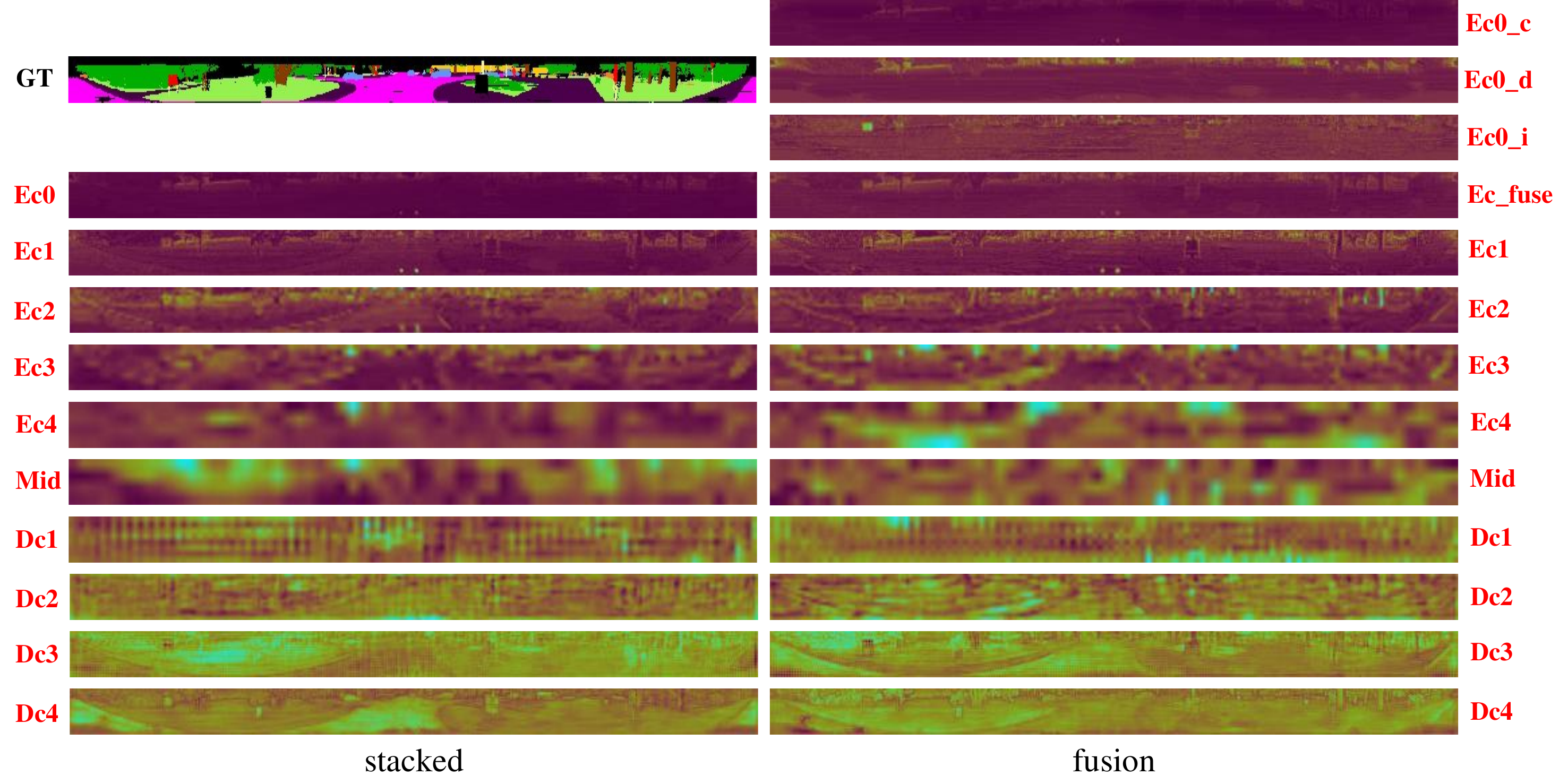}
  \caption{feature maps}
  \end{subfigure}
  \caption{
 For `stacked model' in (a) and `fusion model' in (b), (c) shows the visualization of their feature maps at different building blocks (labeled by red-color text). GT represents the corresponding ground-truth labels for reference.}\label{fig.features}
\end{figure}

Fig. \ref{fig.features}(c) shows visualized feature maps from different building blocks of the ``stacked model” in Fig. \ref{fig.features}(a) and the ``fusion model” in Fig. \ref{fig.features}(b). It can be seen that the three building blocks (Ec0\_c, Ec0\_d, Ec0\_i) learn from the three individual modality, and the fused feature (Ec\_fuse) captures more discriminative information than block Ec0 of the stacked model. This shows that learning modality-specific feature separately helps to capture more representative information in the feature space. This can be further verified by the feature maps of the following building blocks where the fusion model always captures more representative and discriminative features than the stacked model.

\begin{figure}
  \begin{subfigure}[b]{0.24\textwidth}
  \includegraphics[width=\textwidth]{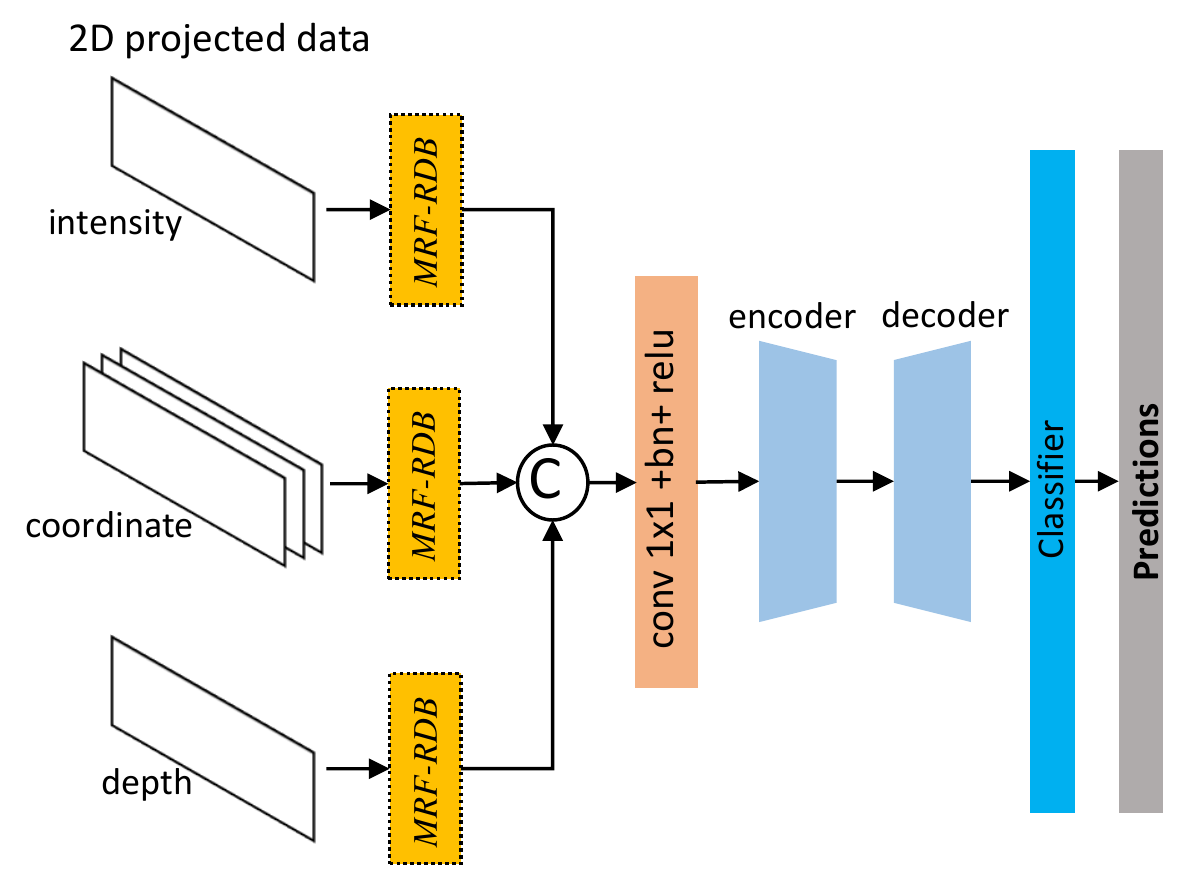}
  \caption{Early-Fusion}
  \end{subfigure}
  \begin{subfigure}[b]{0.24\textwidth}
  \includegraphics[width=\textwidth]{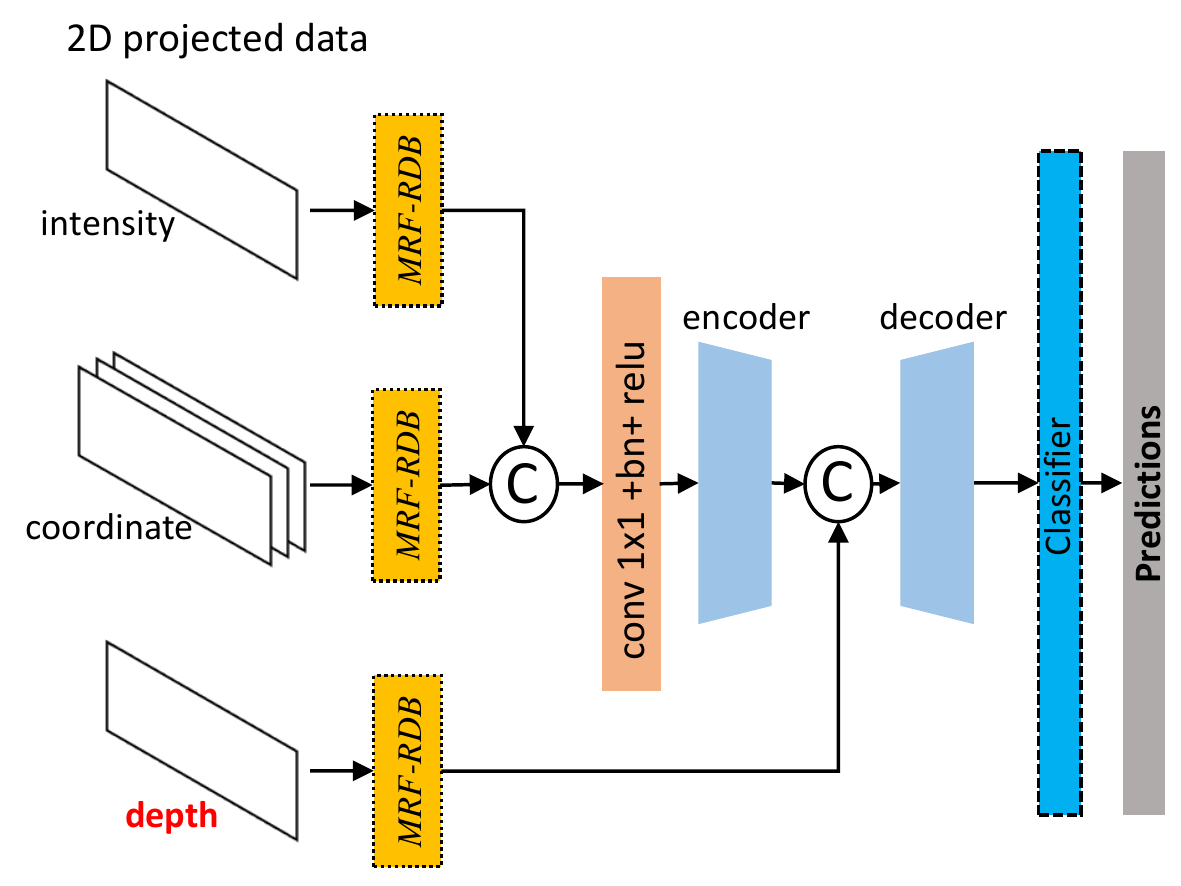}
  \caption{Mid-Fusion \#1}
  \end{subfigure}
  \begin{subfigure}[b]{0.24\textwidth}
  \includegraphics[width=\textwidth]{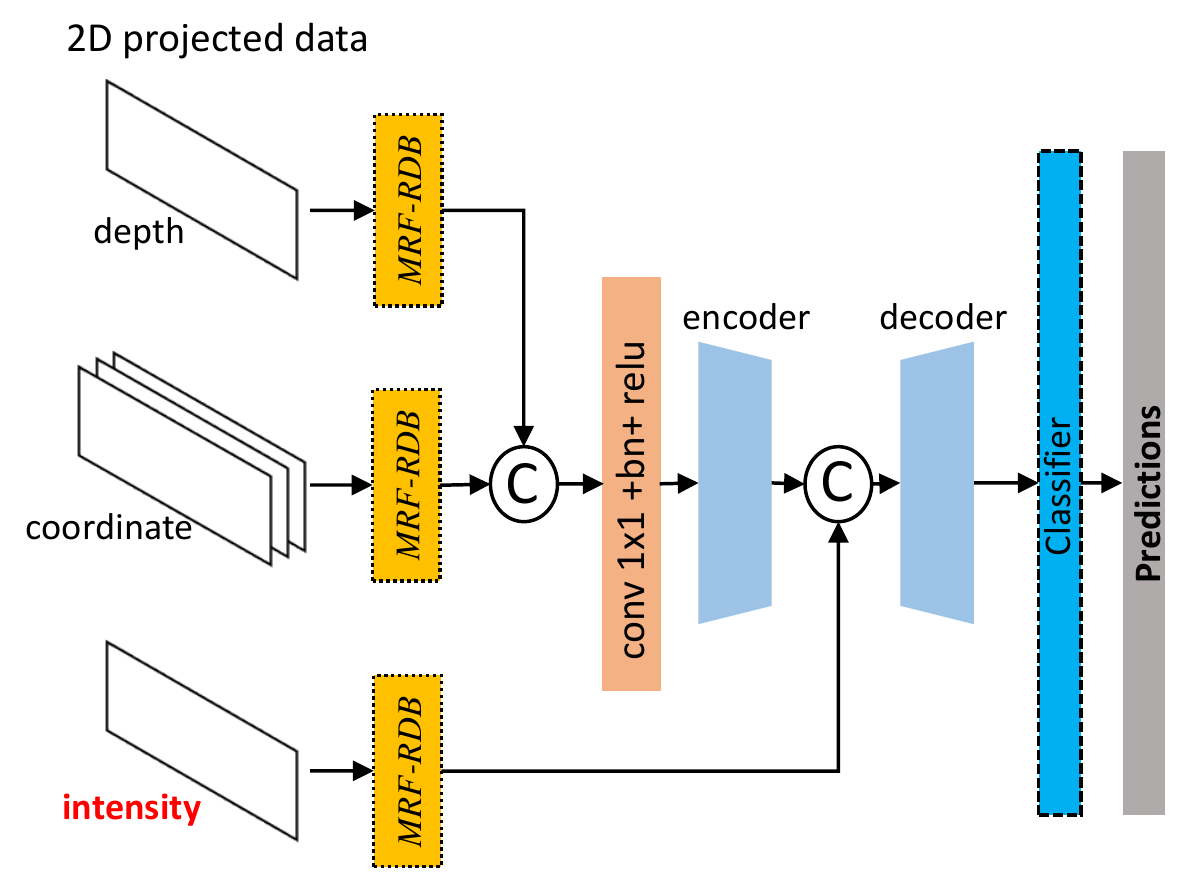}
  \caption{Mid-Fusion \#2}
  \end{subfigure}
  \begin{subfigure}[b]{0.24\textwidth}
  \includegraphics[width=\textwidth]{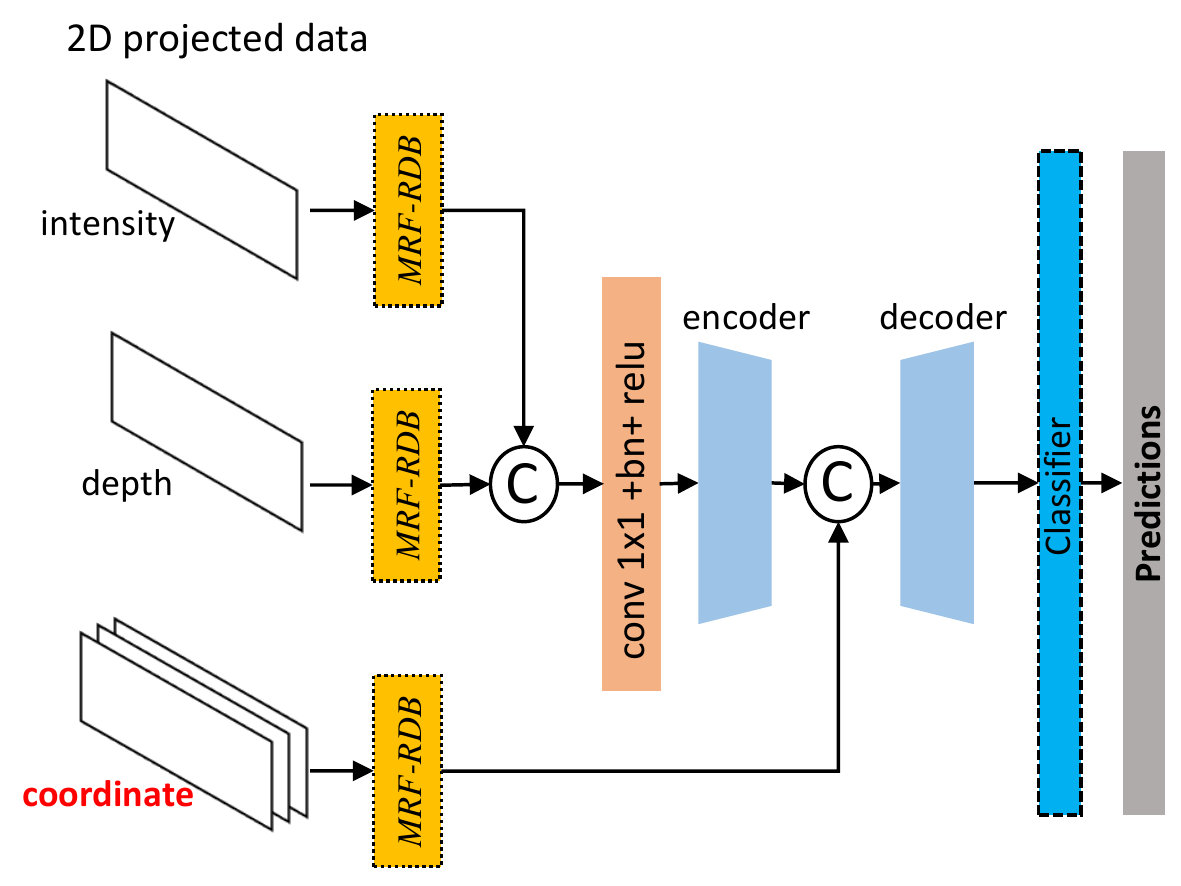}
  \caption{Mid-Fusion \#3}
  \end{subfigure}
  
  \begin{subfigure}[b]{0.24\textwidth}
  \includegraphics[width=\textwidth]{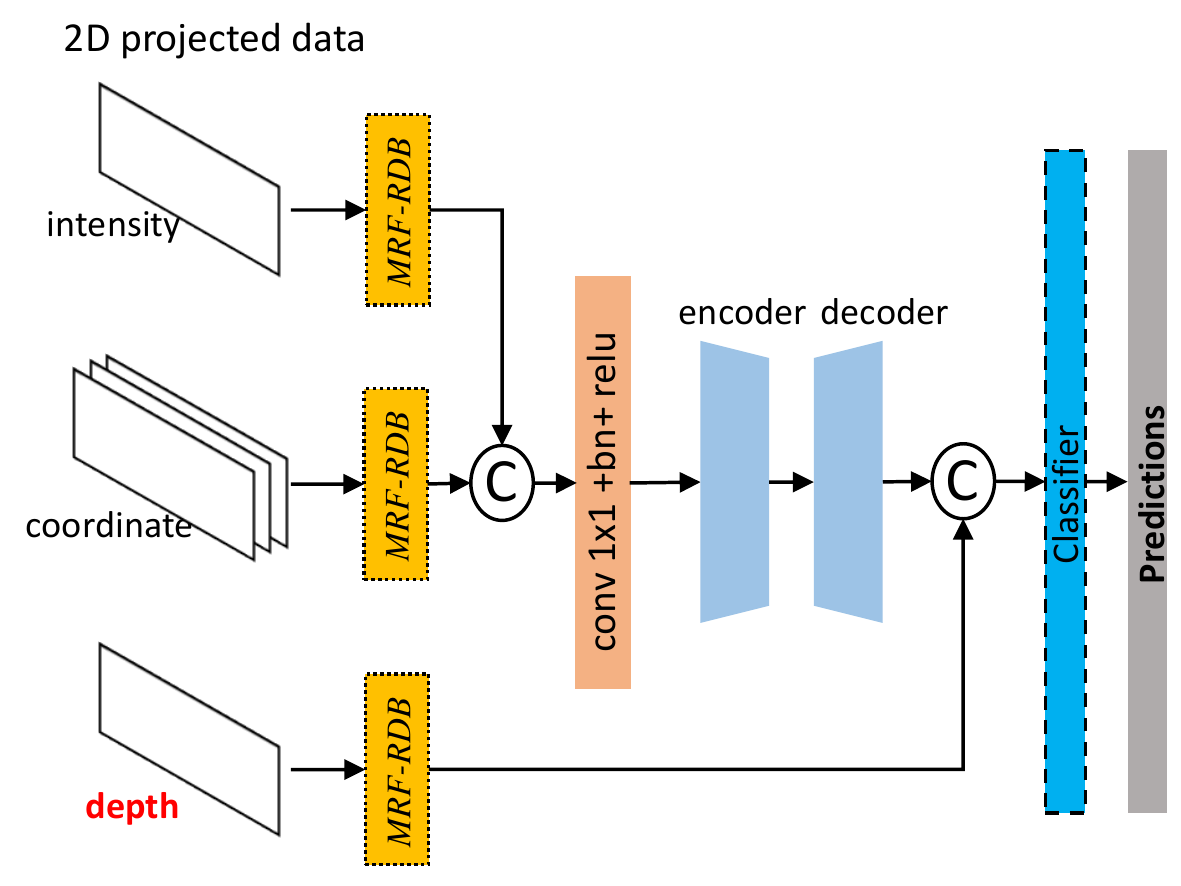}
  \caption{Deep-Fusion \#1}
  \end{subfigure}%
  \begin{subfigure}[b]{0.24\textwidth}
  \includegraphics[width=\textwidth]{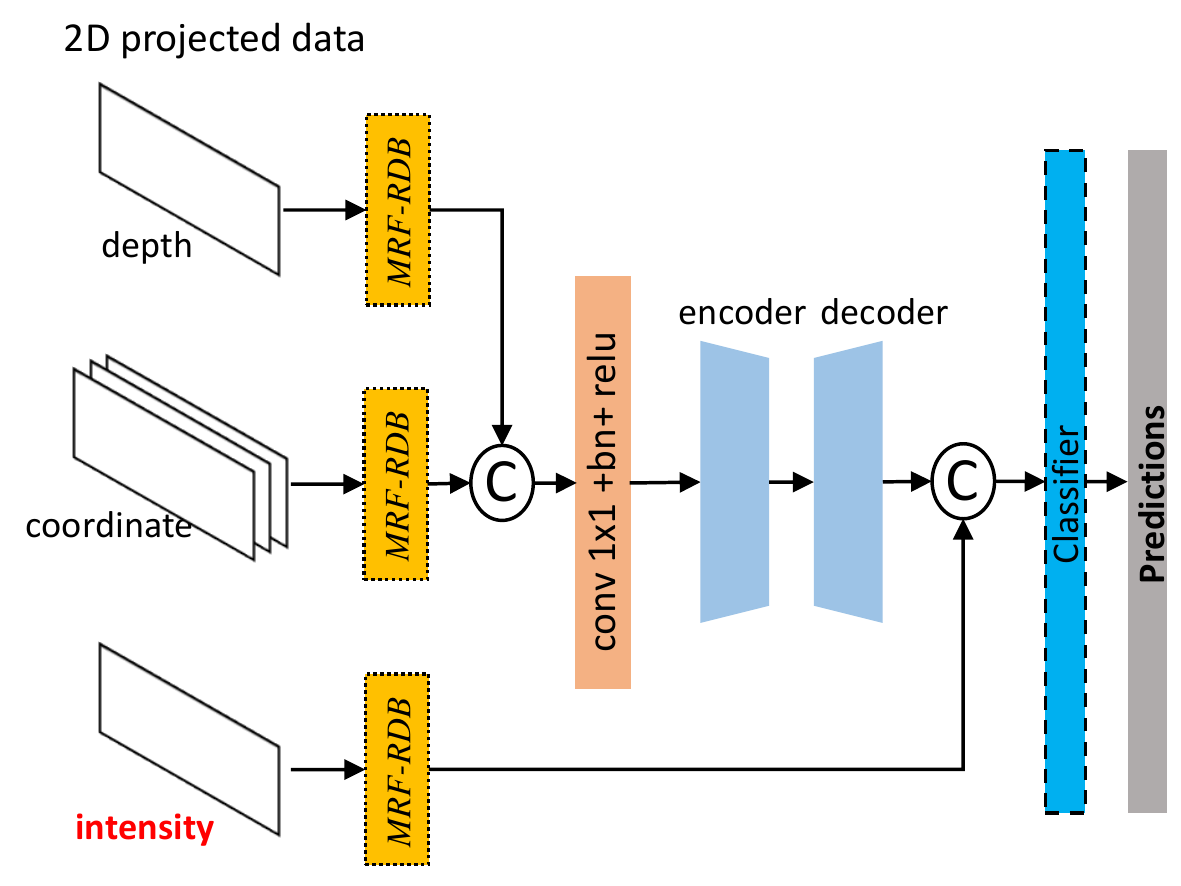}
  \caption{Deep-Fusion \#2}
  \end{subfigure}
  \begin{subfigure}[b]{0.24\textwidth}
  \includegraphics[width=\textwidth]{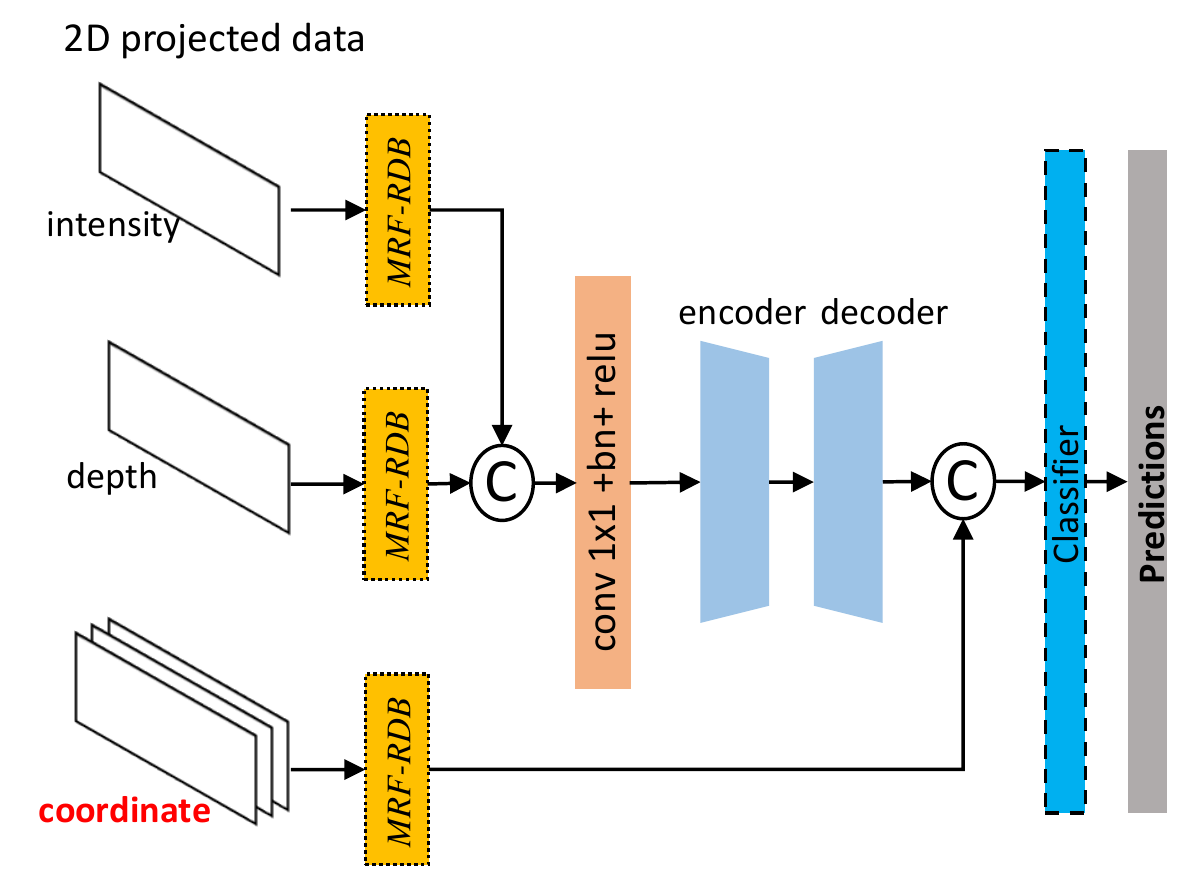}
  \caption{Deep-Fusion \#3}
  \end{subfigure}%
  \caption{Different modality fusion positions.}\label{fig.fuse}
\end{figure}

We also studied the effect of different fusion positions of FPS-Net including early-fusion (fuse before U-Net structure), mid-fusion (fuse before decoder) and deep-fusion (fuse after decoder). 
Fig. \ref{fig.fuse} shows all seven fusion strategies and Table \ref{tab.Ablation-fusion} shows the corresponding mIoU. 
As Table \ref{tab.Ablation-fusion} shows, fusing the three modalities in an early stage achieves the best segmentation while the segmentation deteriorates when the fusing position gets deeper. 
We conjecture that the three modalities of the same point are highly location-aligned and fusing them before the encoder allows the model to learn more representative features in the fusion space. 
In addition, our study shows that fusing intensity before the decoder leads to the worst segmentation performance.

\renewcommand\arraystretch{1.0}
\begin{table}[!h]
  \caption{Ablation study for fusion position, evaluated over the SemanticKITTI validation set. Types are described as Figure \ref{fig.fuse}.} 
  \centering
  \begin{tabular}{ccc}
    \hline
    Type & Position  & mIoU\\
    \hline
    (a) & Early-Fusion & 54.9\\
    (b) & Mid-Fusion 1 & 53.4\\
    (c) & Mid-Fusion 2 & 51.5\\
    (d) & Mid-Fusion 3 & 54.0 \\
    (e) & Deep-Fusion 1 & 53.3\\
    (f) & Deep-Fusion 2 & 53.6\\
    (g) & Deep-Fusion 3 & 53.0 \\
    \hline
  \end{tabular}
  \label{tab.Ablation-fusion}
\end{table}

\subsection{Discussion}

\begin{table}[h]
  \caption{Compatibility experiments about the proposed modality fusion idea (on the SemanticKITTI validation set). Our idea complements with state-of-the-art methods : The three state-of-the-art methods all improve clearly when our modality fusion idea is incorporated as in methods highlighted by `*’. `FPS-Net (w/o fusion)' means using stacked images instead of modality fusion in FPS-Net.}
  \centering
  \begin{tabular}{lccl}
    \hline
    Methods & mIoU & FPS  & \#parameters\\
    \hline
    SqueezeSeg~\cite{wu2018squeezeseg}    & 29.1 & 72.6 & 0.92M \\
    SqueezeSeg* & 30.3 & 52.0 & 1.03M (+17.7\%)\\
    \hline
    SqueezeSegV2~\cite{wu2019squeezesegv2} & 39.2 & 62.8 & 0.92M\\
    SqueezeSegV2* & 40.5 & 44.9 & 1.09M (+17.5\%)\\
    \hline
    RangeNet++~\cite{milioto2019rangenet++} & 48.2 & 14.9 & 50.38M \\
    RangeNet++*   & 49.1 & 14.1 & 50.41M (+0.06\%)\\
    \hline
    FPS-Net (w/o fusion) & 52.3 & 22.1 & 55.56M \\
    FPS-Net  & 54.9 & 20.8 & 55.73M (+0.29\%)\\
    \hline
  \end{tabular}
  \label{tab.extend_model}
\end{table}

We also studies whether the proposed separate learning and fusion approach can complement state-of-the-art point cloud segmentation methods which stack all projected channel images as a single input in learning. The compared methods include SqueezeSeg \cite{wu2018squeezeseg}, SqueezeSegV2 \cite{wu2019squeezesegv2} and RangeNet++ \cite{milioto2019rangenet++} that all stacked five channel images as a single input in learning. In the experiments, we changed their networks by inputting channel images of the three modality separately (together with early-fusion) and trained segmentation model over the dataset sub-SemanticKITTI. Table \ref{tab.extend_model} shows experimental results on the SemanticKITTI validation set. As Table \ref{tab.extend_model} shows, all three adapted networks (with separate learning and fusion) outperform the original networks (with stacked learning) clearly. 
Note that the parameter increases are relatively large for SqueezeSeg and SqueezeSegV2 which are pioneer lightweight networks. But for recent models including RangeNet++ and our FPS-Net, the proposed fusion approach introduces very limited extra parameters (less than 0.3\%) but significant improvement in segmentation. This further verifies the constraint of modality gaps in semantic segmentation of point cloud. More importantly, it demonstrates that our proposed separate learning and fusion idea can be incorporated into most existing techniques with consistent performance improvements.

\section{Conclusion}\label{Sec.conclusion}
This paper identifies the modality gap problem existed in spherical projected images ignored by previous methods and designed a separate learning and fusing architecture for it. Based on this idea, it proposes an end-to-end convolutional neural network named \textit{FPS-Net} for semantic segmentation on large scale LiDAR point cloud, which processes the projected LiDAR scans fast and efficiently. Experiments prove the effectiveness of our idea and show that our model achieves better performances in contrast to recently published models in the SemanticKITTI benchmark and a significant improvement in the KITTI benchmark with comparable computation speed. Extensive experiments show that our idea is also contributory to classical spherical projection-based methods and is able to improve their performances.

\section*{ACKNOWLEDGEMENTS}\label{ACKNOWLEDGEMENTS}
This research was conducted at Singtel Cognitive and Artificial Intelligence Lab for Enterprises (SCALE@NTU), which is a collaboration between Singapore Telecommunications Limited (Singtel) and Nanyang Technological University (NTU) that is funded by the Singapore Government through the Industry Alignment Fund ‐ Industry Collaboration Projects Grant.

\bibliography{mybibfile}

\end{document}